\documentclass[10pt,journal]{IEEEtran}
\usepackage{amsmath,epsfig}
\usepackage{multirow}
\usepackage{float}
\usepackage{color}
\usepackage{pgfplotstable}
\usepackage{booktabs}
\usepackage{todonotes}
\usepackage{graphicx} 
\usepackage{breakurl}
\usepackage{caption}
\usepackage{tikz}
\usepackage{xcolor}
\usepackage{epstopdf}
\usepackage{amssymb}
\usepackage{soul,xcolor}
\usepackage{siunitx}
\usepackage{pifont}
\usepackage[switch]{lineno} 
\usepackage{lipsum} 
\usepackage{multirow}
\usepackage{epstopdf}
\AppendGraphicsExtensions{.tif}
\usepackage{CJKutf8}

\begin{document}
\sloppy  

\title{UAVStereo: A Multiple Resolution Dataset for Stereo Matching in UAV Scenarios}

\author{Xiaoyi Zhang\IEEEauthorrefmark{1}, Xuefeng Cao, Anzhu Yu, Wenshuai Yu, Zhenqi Li,  Yujun Quan
\IEEEcompsocitemizethanks{
\IEEEcompsocthanksitem {
}

\IEEEcompsocthanksitem{
X. Zhang, X. Cao, A. Yu, Z. Li, Y. Quan are with the PLA Strategic Support Force Information Engineering University, Zhengzhou 450001, China. W.Yu is with the College of Civil and Transportation Engineering, Shenzhen University, Shenzhen 518060, China. Corresponding author: Xuefeng Cao (CAO\_Xue\_Feng@163.com)}
}
}
 
\IEEEtitleabstractindextext{%
\begin{abstract}
Stereo matching is a fundamental task for 3D scene reconstruction. Recently, deep learning based methods have proven effective on some benchmark datasets, such as KITTI and Scene Flow. UAVs (Unmanned Aerial Vehicles) are commonly utilized for surface observation, and their captured images are frequently used for detailed 3D reconstruction due to high resolution and low-altitude acquisition. At present, the mainstream supervised learning network requires a significant amount of training data with ground-truth labels to learn model parameters. 
However, due to the scarcity of UAV stereo matching datasets, the learning-based network cannot be applied to UAV images. 
To facilitate further research, this paper proposes a novel pipeline to generate accurate and dense disparity maps using detailed meshes reconstructed by UAV images and LiDAR point clouds. Through the proposed pipeline, this paper constructs a multi-resolution UAV scenario dataset, called UAVStereo, with over 34k stereo image pairs covering 3 typical scenes. 
As far as we know, UAVStereo is the first stereo matching dataset of UAV low-altitude scenarios. The dataset includes synthetic and real stereo pairs to enable generalization from the synthetic domain to the real domain. Furthermore, our UAVStereo dataset provides multi-resolution and multi-scene images pairs to accommodate a variety of sensors and environments. In this paper, we evaluate traditional and state-of-the-art deep learning methods, highlighting their limitations in addressing challenges in UAV scenarios and offering suggestions for future research. The dataset is available at https://github.com/rebecca0011/UAVStereo.git

\end{abstract}

}

\maketitle%
\IEEEdisplaynontitleabstractindextext
\IEEEpeerreviewmaketitle
\maketitle

\ifCLASSOPTIONcompsoc
    \IEEEraisesectionheading{\section{Introduction}\label{sec:introduction}}
\else
    \section{Introduction}
    \label{sec:intro}
\fi

\IEEEPARstart{O}{ne} of the most active areas of research in photogrammetry and computer vision is three-dimensional (3D) reconstruction of the environment via dense matching, which can be performed in stereo (in two views) \cite{scharstein2002taxonomy} or multi-view stereo (MVS) \cite{jensen2014large}. Among image-based approaches, stereo matching \cite{poggi2021synergies}, where expected correspondences are on the epipolar lines, is arguably the most popular and intensively researched technique. Significant progress in this field has been made in terms of accuracy and cross-domain performance. Thanks to stereo benchmarks \cite{geiger2012we}\cite{yang2019drivingstereo}\cite{schops2017multi}\cite{scharstein2014high}, researchers have achieved high accuracy on benchmarks of driving scenarios and indoor environments. Furthermore, some aerial stereo datasets paved the way for deep learning to succeed also in aerial stereo images \cite{wu2021new}\cite{cournet2020ground}\cite{liu2020novel}. However, lack of large-scale datasets hinders research on cross-domain performance and application of the stereo matching algorithm to UAV images.

UAV introduces a low-cost alternative to classical aerial photogrammetry for large-scale topographic mapping or detailed 3D recording of ground information and is a valid complementary solution to terrestrial observation. With such UAV images, current networks face at least three challenges:
\begin{itemize}
\item Larger disparity search space: The conversion between disparity and depth can be written as $disparity=\frac{Bf}{Depth}$, with baseline $B$ and focal length in pixels $f$. Focal length in pixels $f$ can be further expressed as $\frac{W_{px} × f_{mm}}{W_{CCD}}$, in which $W_{px}$, $f_{mm}$, ${W_{CCD}}$ represents image width in pixels, focal length in mm, CCD width in mm respectively. With the advancement of digital cameras, the resolution of acquired images is increasing, resulting in a larger disparity search space, which places additional performance requirements on the algorithm. 
\item Bigger possibility of ill-areas: UAV are often used to capture information about the ground surface, such as forests and grasslands, where features are harder to match. In addition to the low-altitude acquisition characteristics, it is easier for UAVs to acquire images containing ill-areas such as textureless and repetitive textures, which is extremely challenging and may in many cases be an inherently ill-posed problem.
\item More varied disparity distribution: the UAV's lightweight and adaptable characteristics allow it to collect images from a variety of heights, where disparity distribution differs greatly from driving and indoor scenarios and is significantly more variable.The majority of current algorithms are applied to datasets with the same disparity distribution, such as driving and indoor, which presents an additional challenge.
\end{itemize}

In addition to the above properties of UAV images, the literature \cite{yang2019drivingstereo} points out that the existing algorithm has a huge gap between the synthetic domain and the real domain. To shorten this gap, synthetic image pairs are used in advance for pretraining and a small amount of real images are used therewith for finetuning, which can significantly improve the pretrained models' capacity on real data. Obviously, current single synthetic dataset or the real dataset cannot match the requirements. It is necessary to combine synthetic and real data in one dataset.

Towards this goal, we propose a novel UAV scenarios dataset that co-exists with synthetic and real data. For synthetic data, we propose a large-scale stereo dataset with sufficient variation, realism, and size to successfully train large networks in UAV scenarios. For real data, the acquired images are provided after four-step processing (including initial disparity maps generation, epipolar images generation, epipolar disparity maps generation, and post-processing). In addition, the original resolution stereo pairs and corresponding disparity maps are supplied for high-resolution network evaluation. The main contributions of our paper are:
\begin{itemize}
\item We propose a pipeline (detail in \ref{Synthetic Dataset} and \ref{real Dataset}) capable of generating dense disparity for both synthetic and real images from UAV-obtained images and point clouds
\item We construct a new UAVStereo, which consists of image pairs and dense disparity maps for three representative UAV scenes. 
To shorten the gap between synthetic and real domain, we construct both synthetic and real data to increase the availability of in real scenarios and decrease the quantity demand for real data. In order to adapt to the imaging characteristics and disparity distribution in UAV scenarios, we published multi-resolution images and corresponding disparity maps.
\item We evaluate traditional and state-of-the-art deep techniques on our dataset. Experimental results across different datasets and stereo methods demonstrate that our dataset is more suitable for the UAV scenario. Our dataset also presents challenges to current algorithms in terms of resolution, disparity search range and geospatial feature matching.
\end{itemize}

\section{Related work}
\label{sec:Related work}
\subsection{Stereo Matching Methods} 
For many years, most algorithms have been solving stereo matching problem following a typical four-step pipeline\cite{scharstein2002taxonomy}: Matching cost computation, cost aggregation, disparity computation, and disparity refinement. Among the vast literature on traditional algorithms \cite{de2011linear} \cite{di2004fast} \cite{hosni2012fast} \cite{yang2012non} \cite{yoon2006adaptive}, Semi-Global Matching (SGM) \cite{hirschmuller2007stereo} is the most popular, which is a reference approach combining mutual information and dynamic programming optimization on several directions \cite{hirschmuller2005accurate}.

With the development of deep learning, the first research efforts focused on replacing the individual steps of the conventional pipeline with deep learning counterparts. For instance, 2D convolutional neural networks (CNN) prove effective in feature extraction \cite{zbontar2016stereo}. SGM-Net uses a CNN network to provide learned penalties for SGM \cite{seki2017sgm}.

Then, end-to-end deep stereo networks rapidly gained the main stage \cite{kendall2017end}\cite{mayer2016large}\cite{pang2017cascade}. Inspired by FCN used in semantic segmentation \cite{long2015fully}\cite{chen2017deeplab}, DispNet \cite{mayer2016large} adopts an encoder-decoder architecture to enable end-to-end disparity regression, where the matching cost can be directly integrated to encoder volumes. GC-Net \cite{kendall2017end} combines contextual information by 3D convolutions over a cost volume. PSMNet \cite{chang2018pyramid} integrates global context information using spatial pyramid pooling and regularizes cost volume using stacked multiple hourglass 3D convolutional networks. To tackle the high memory consumption for high-resolution image matching, Deepprunner \cite{duggal2019deeppruner} proposes to prune the 3D cost volume with a differential patch match method. STTR \cite{li2021revisiting} revisits the problem from a sequence-to-sequence correspondence perspective and replaces cost volume construction with dense pixel matching using position information and attention.

While these methods were developed by the computer vision community on the indoor or driving dataset, numerous researchers introduced stereo matching networks into geospatial aerial images and proved to be effective \cite{knobelreiter2018self}\cite{he2022hmsm}.

Compared with the traditional method, learning-based stereo matching networks have shown excellent feature matching capabilities in many scenarios and have been applied in aerial image stereo matching due to their superior cross-domain generation capabilities. However, the capability of the network in UAV imagery is not validated due to the lack of data.

\subsection{Stereo benchmarks}
Among the factors behind the rapid development of stereo matching, the growing availability of datasets plays a crucial role. Table \ref{tab:Dataset} lists some datasets for stereo matching proposed by the aerial photogrammetry and computer vision communities. Some are for driving scenes: the KITTI datasets in two versions, KITTI 2012 \cite{geiger2012we} and KITTI 2015 \cite{menze2015object}; the large-scale stereo DrivingStereo \cite{yang2019drivingstereo} and AppolloScape \cite{huang2019apolloscape}. Then Middlebury 2014 \cite{scharstein2014high} framing indoor environments, and ETH3D \cite{schops2017multi} including both indoor and outdoor scenes are also popular and widely utilized. In aerial photogrammetry, SatStereo \cite{patil2019new}, \cite{bosch2016multiple} and ISPRS2019Benchmark \cite{wu2021new} dataset is frequently used for dense matching evaluation. Using advanced computer graphics, The Scene Flow \cite{mayer2016large}, Virtual KITTI \cite{gaidon2016virtual} , and MPI Sintel \cite{butler2012naturalistic} datasets synthesize dense disparity maps, yet it remains a huge gap between the synthetic domain and the real world.

UAVs are commonly used for earth observation because of their mobility, flexibility and low cost. However, UAV images processing requires high time and computational memory, due to its high resolution. We are attempting to reduce the processing time by incorporating stereo matching network into the it. To this goal, large-scale UAV scenarios datasets are required to train the network. In this paper, we propose the UAVStereo dataset, which contains a large amount of image pairs with dense disparity to facilitate the training and testing of stereo matching networks.

\begin{table*}[h]
  \centering
  \caption{Comparison of available stereo datasets.}
    \begin{tabular}{llllll}
    \toprule
   Dataset & Year & Scenario & Stereo number & Resolution & Disparity density \\
   \midrule
   KITTI2012 \cite{geiger2012we} & 2012 & Driving & 389 & 1226×370 & sparse \\
   KITTI2015 \cite{menze2015object} & 2015 & Driving & 400 & 1242×375 & sparse \\
   DrivingStereo \cite{yang2019drivingstereo} & 2019 & Driving & 182188 & 1762×800 & sparse \\
   AppolloScape \cite{huang2019apolloscape} & 2019 & Driving & 19035 & 3384×2710 & sparse \\
   ETH3D \cite{schops2017multi} & 2017 & Indoor + Outdoor & 47 & 940×490 & dense \\
   SatStereo \cite{patil2019new} & 2019 & Aerial & 72 & 1298×1286 & dense \\
   ISPRS2019Benchmark \cite{wu2021new} & 2021 & Aerial & 1092 & 1024×1024 & sparse \\
   WHUStereo \cite{liu2020novel} & 2020 & Aerial & 10979 & 768×384 & sparse \\
   MPI Sintel \cite{butler2012naturalistic} & 2012 & Synthetic animation & 564 & 1024×436 & dense \\
   Scene Flow \cite{mayer2016large} & 2016 & Synthetic animation & 26066 & 960×540 & dense \\
   Virtual KITTI \cite{gaidon2016virtual} & 2016 & Synthetic driving & 21260 & 1242×375 & dense \\
   \textbf{UAVStereo(Ours)} & \textbf{2022} & \textbf{Synthetic and real UAV} & \textbf{38781} & \textbf{Multiple Resolution} & \textbf{dense} \\
   \bottomrule
   \end{tabular}%
     \label{tab:Dataset}%
\end{table*}%

\section{UAVStereo Dataset}
This section introduces the UAVStereo data production process. 
The data acquisition system and covering areas in section \ref{Data Acquisition}. The data production pipelines for synthesis and real data are described in \ref{Synthetic Dataset} and \ref{real Dataset} respectively. 

\begin{figure*}[!t]
	\centering
	\includegraphics[width=0.9\textwidth]{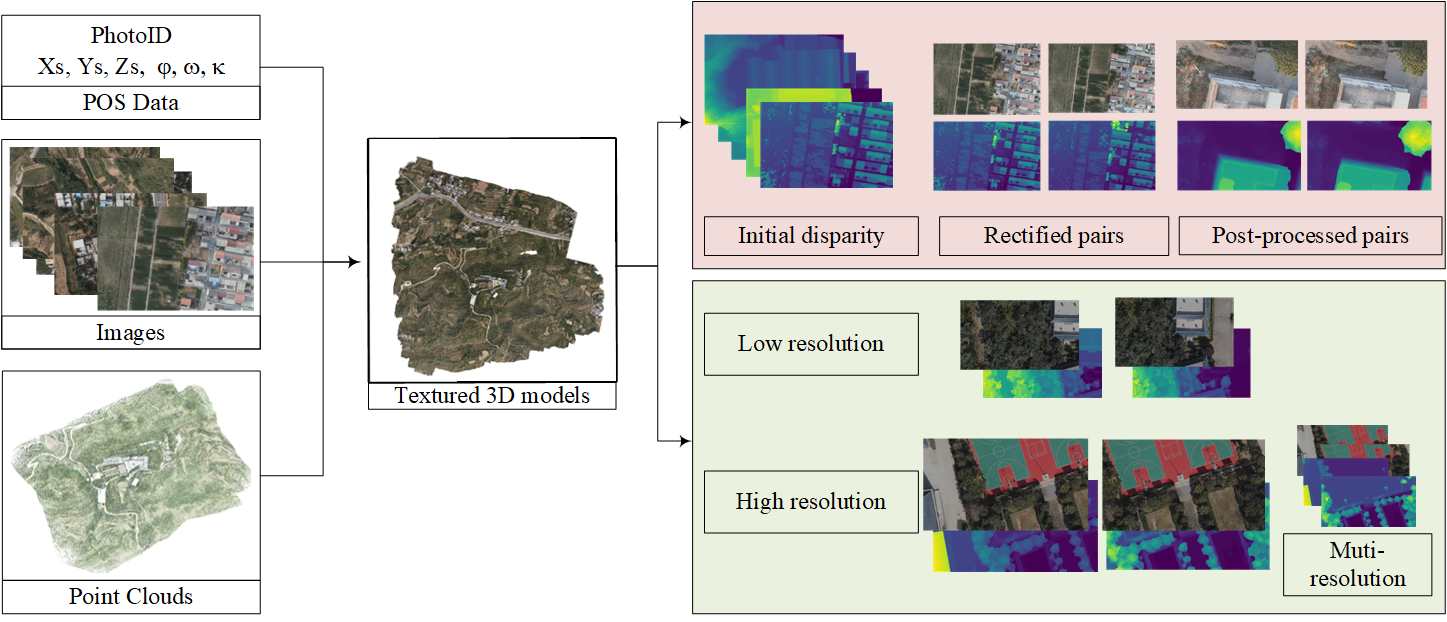}
	\caption{Pipeline of data generation. The red box shows the real data generation process. The green box shows the synthetic data rendering process.
	}
	\label{fig:Dataset generation pipeline}
\end{figure*}

\subsection{Data Acquisition}
\label{Data Acquisition}
The commercial unmanned aerial vehicle DJI Matrice 300 is a widely-used platform for earth observation. We chose DJI Matrice 300 as the platform, which was equipped with LiDAR sensor Zenmuse L1 and full-frame imaging sensor Zenmuse P1, obtaining images and point clouds respectively, in the designated areas. The imaging sensor Zenmuse P1 has a 35.9 $mm$ × 24 $mm$ full-frame sensor at pixel size of 4.4 $\mu$$m$, allowing to capture high-quality photographs with a resolution of 8192 x 5460 $px$. Zenmuse L1 integrates a Livox LiDAR module and a camera, allowing it to capture the details of complex structures and generate true-color point cloud models. The horizontal accuracy and vertical Accuracy of L1 radar are 10 $cm$ and 5 $cm$ respectively. The maximum range of DJI L1 is 190$m$ at 10\%, 100$klx$ and 450$m$ at 80\%, 0$klx$.

The point clouds acquired by Zenmuse L1 LiDAR is first converted into the standard las format by DJI Terra \footnote{https://www.dji.com/au/dji-terra}. Then 3D digital surface models with OBJ format were reconstructed using Daspatial GET3D Cluster software \footnote{https://www.daspatial.com/cn/gcluster} from a substantial amount of images and point clouds. To make the surface model more accurate to the actual scene, we manually corrected the photos with severe rotation and the points with large errors in the aerotriangulation results.

\begin{figure*}[!tbh]
\small
   \centering
		\newcommand{\tabincell}[2]{\begin{tabular}{@{}#1@{}}#2\end{tabular}}
		\centering
		\resizebox{\textwidth}{!}{
		\begin{tabular}
{m{5.0cm}<{\centering}m{5.0cm}<{\centering}m{5.0cm}<{\centering}}

\includegraphics[width=0.3\textwidth]{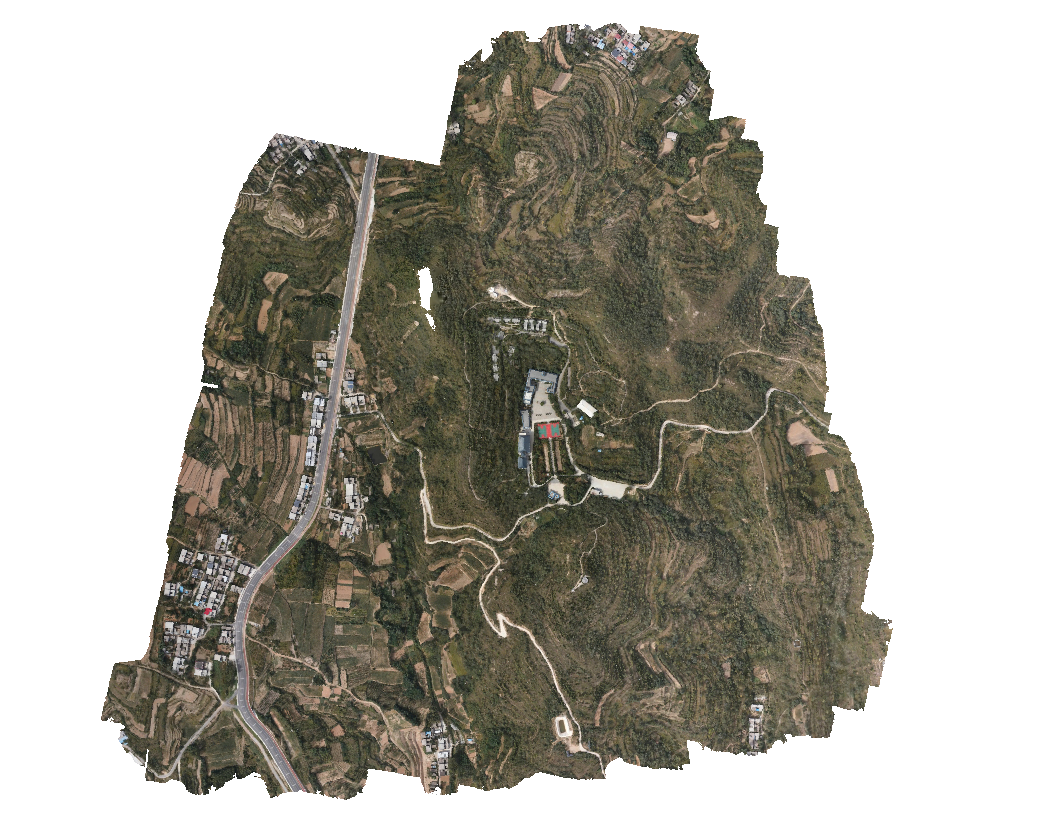}
			& 
\includegraphics[width=0.3\textwidth]{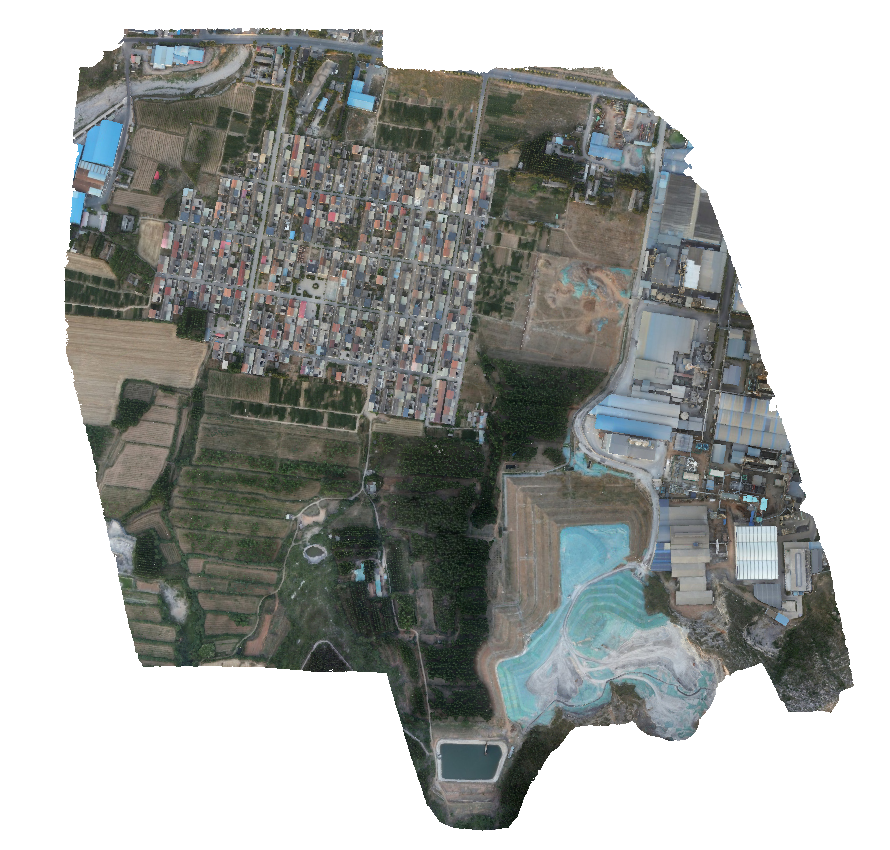}
			& 
\includegraphics[width=0.3\textwidth]{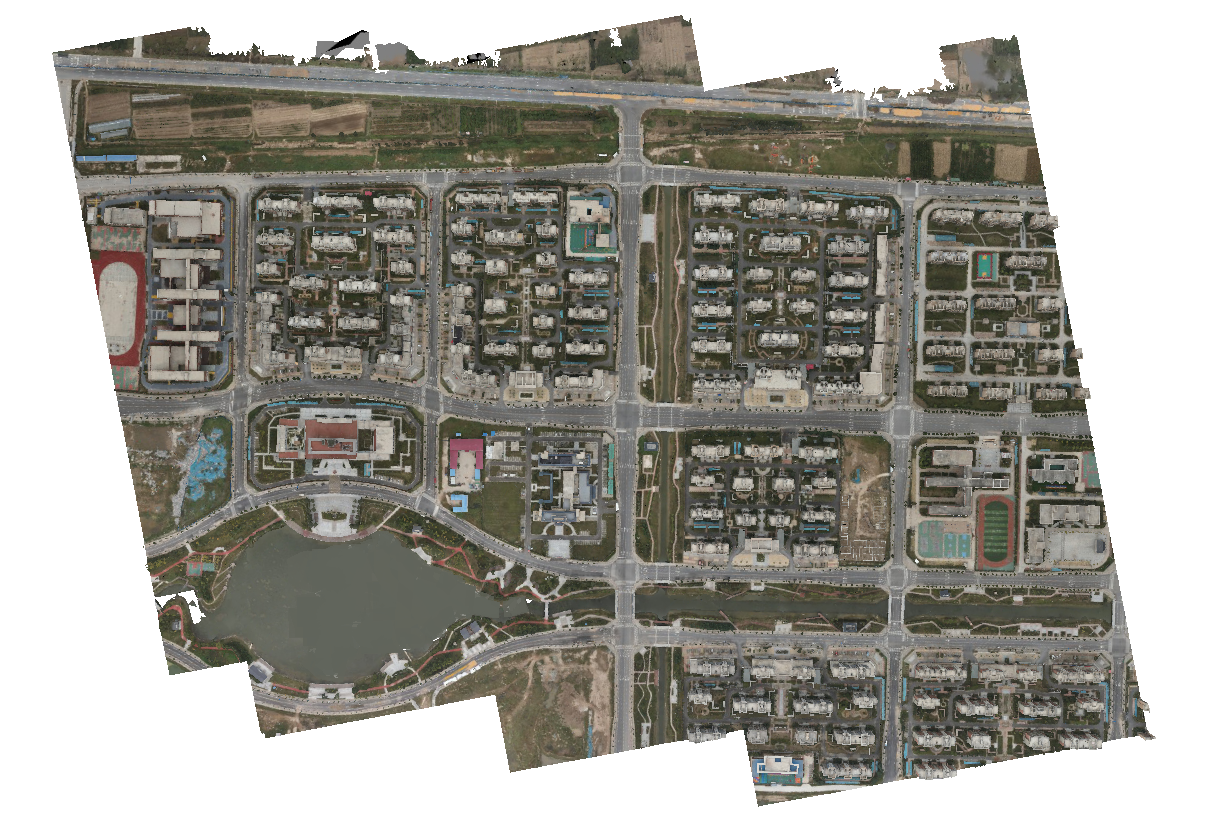}
             \\
(a) forest area & (b) mining area & (c) residential land
\\
\caption{Data acquisition areas.}
\label{fig:Scenarios}
\end{tabular} }  %
\end{figure*}

Three different scenarios are included in our UAVStereo, including residential land, forest and mining area. As shown in Fig. \ref{fig:Scenarios}, the residential area contains dense and regular tall buildings, flat roadways and other urban scene features, covering about 700 × 1200 $m^{2}$. This area provides an urban scene with disparity saltation like building. 
The forest area contains high-coverage trees, several houses and other field scene components, covering about 1350 × 1500 $m^{2}$. This region has textureless and repeated textured images, which presents a difficulty for stereo matching algorithms.
The mining zone is made up of around 700 × 700 $m^{2}$ of agriculture, low structures, and bare ground, which contains continuous variation of disparity.
These three areas are representative areas for UAV earth observation and can represent different disparity distribution.

\subsection{Synthetic Dataset}
\label{Synthetic Dataset}
The model trained on synthesized data can provide a good initial pretrained model for the application in real UAV images. Therefore, we generate multi-resolution and multi-scene UAV data to adapt to the application of different sensors and different scenes. Similar to Scene Flow \cite{mayer2016large}, we used the open-source 3D creation suite Blender \footnote{https://www.blender.org/} to simulate the flight path of drones and render the results into tens of thousands of frames. As shown in fig \ref{fig:Dataset generation pipeline}, We rendered textured 3D models into color images and the corresponding ground truth disparity maps, generating a synthetic dataset subset consisting of both low and high resolution subsets.

Given the intrinsic camera parameters (focal length $f$, principal point ($x_{0}, y_{0}$)), the render settings (image size $W, H$, and sensor size and format) and the exterior orientation (camera center $(Xs, Ys, Zs)$ and three rotational angles $(\varphi, \omega, \kappa)$, baseline $B$ ), Blender can directly retrieved the depth of each pixel from imported OBJ models. 
We generated stereo images using Blender‘s Stereoscopy following the designed drone flight path. According the formula $disparity=\frac{Bf}{Depth}$, we adjusted the render settings and converted the depth to disparity using the known configuration of the virtual stereo rig, generating the corresponding disparity maps 

Adjusting the external orientation elements, the synthetic images and corresponding disparity maps are acquired at 100 - 300 $m$ above the model with high overlap. For all frames and views, we provide 8-bit RGB images and disparity maps with lossless pfm format. We rendered all image data using a virtual focal length of 35 $mm$ on a 36 $mm$ wide simulated sensor. We released the high and low resolution subsets at 960 × 540 $px$ (same as Scene Flow) and 8192 × 5460 $px$ (same as Zenmuse P1). At the same time, we also resize high-resolution images to 3840 × 2160 $px$ and 1920 × 1080 $px$ for multi-resolution evaluation. The baseline is set to 1 $m$ to 15 $m$ for the low-resolution subset due to the image size limitation, while 15 $m$ to 35 $m$ for the high-resolution subset. The image size, camera center and baseline length of these two subsets are greatly different, which can not only provide multi-resolution images, but also test the robust performance of stereo matching algorithms. 

In Tab. \ref{tab:samplesSynImages}, we present the sample data in Synthetic subset with a baseline length of 15 $m$.

\begin{table*}[h]
\small

   \centering
		\newcommand{\tabincell}[2]{\begin{tabular}{@{}#1@{}}#2\end{tabular}}
		\centering
		\resizebox{\textwidth}{!}{
		\begin{tabular}{m{1.0cm}<{\centering}m{4.5cm}<{\centering}m{4.5cm}<{\centering}m{4.5cm}}

    {\rotatebox{90}{Left image}}	&
\includegraphics[width=0.26\textwidth]{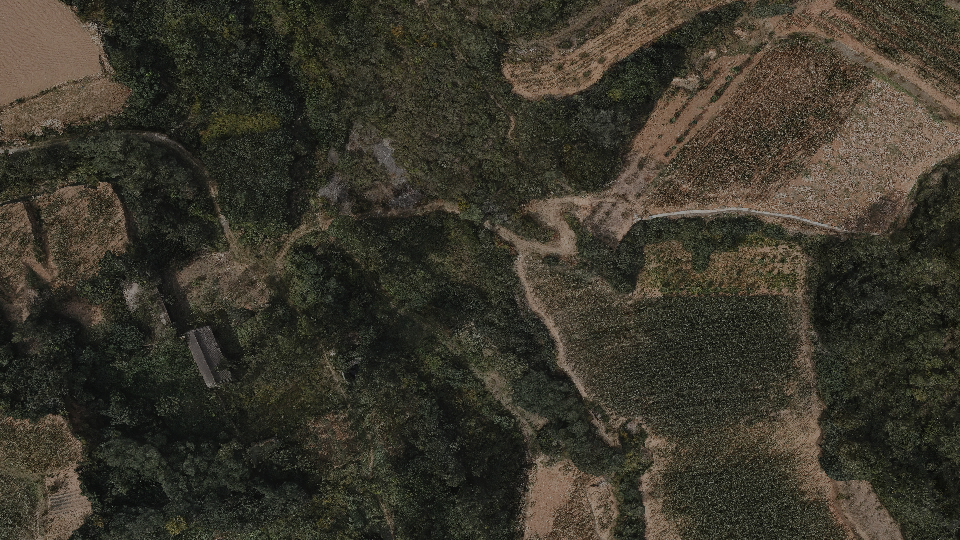}
			& 
\includegraphics[width=0.26\textwidth]{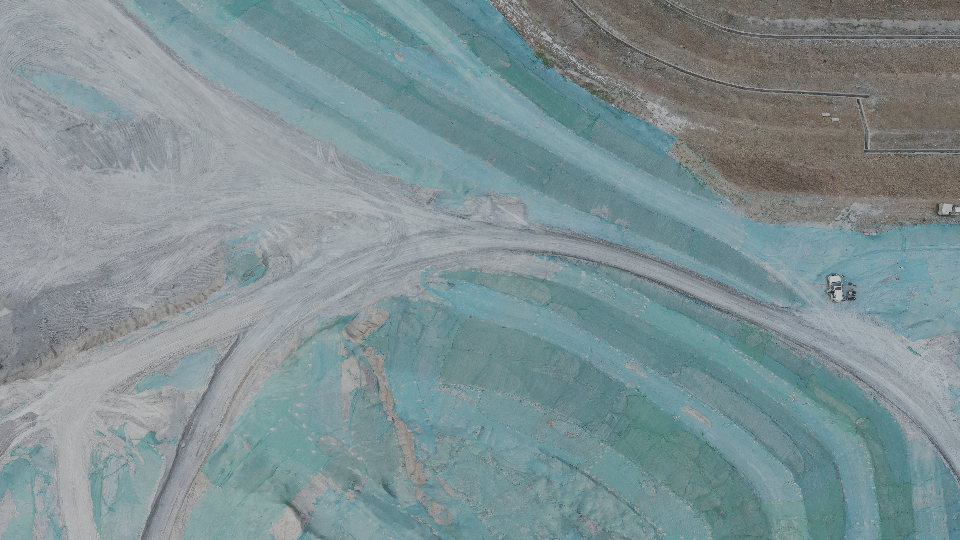}
			& 
\includegraphics[width=0.26\textwidth]{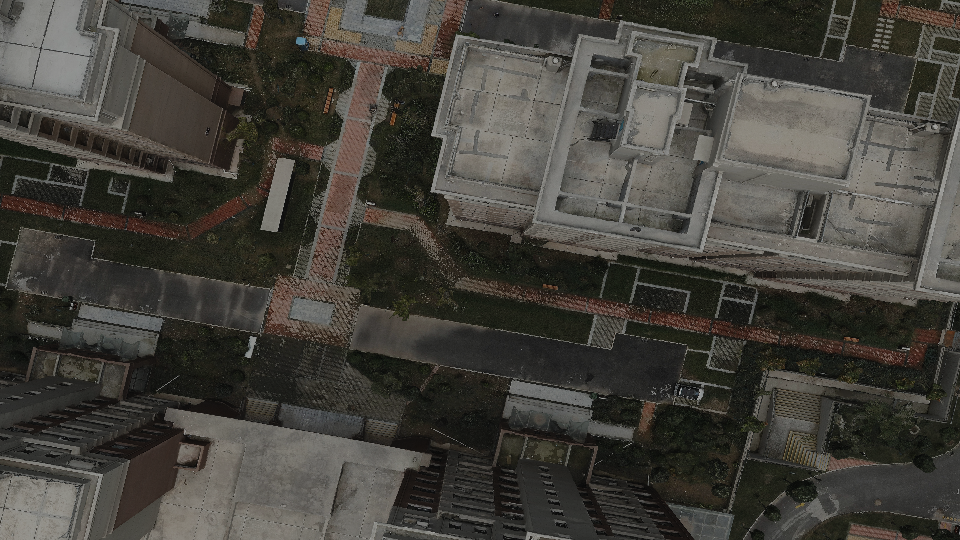}
		\\ 
	{\rotatebox{90}{Right image}}	&
\includegraphics[width=0.26\textwidth]{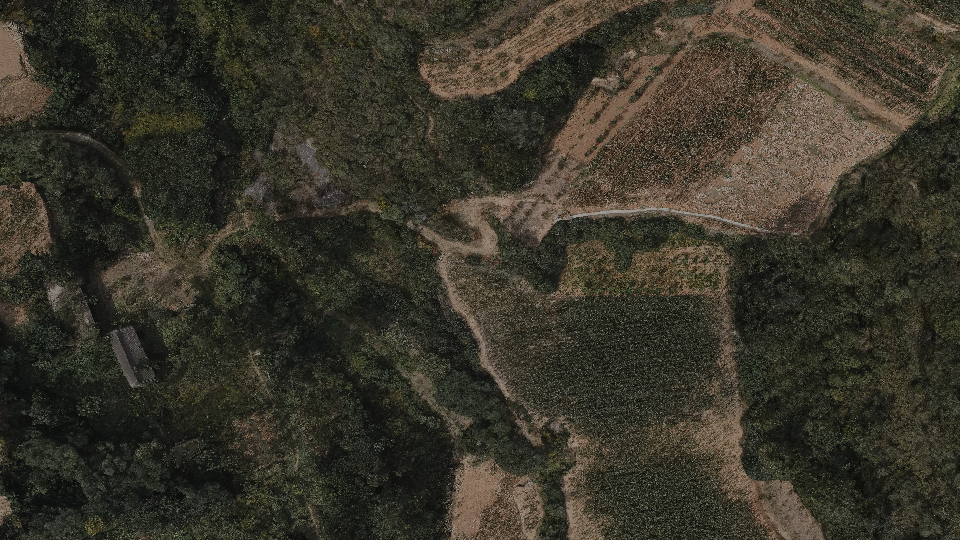}
			& 
\includegraphics[width=0.26\textwidth]{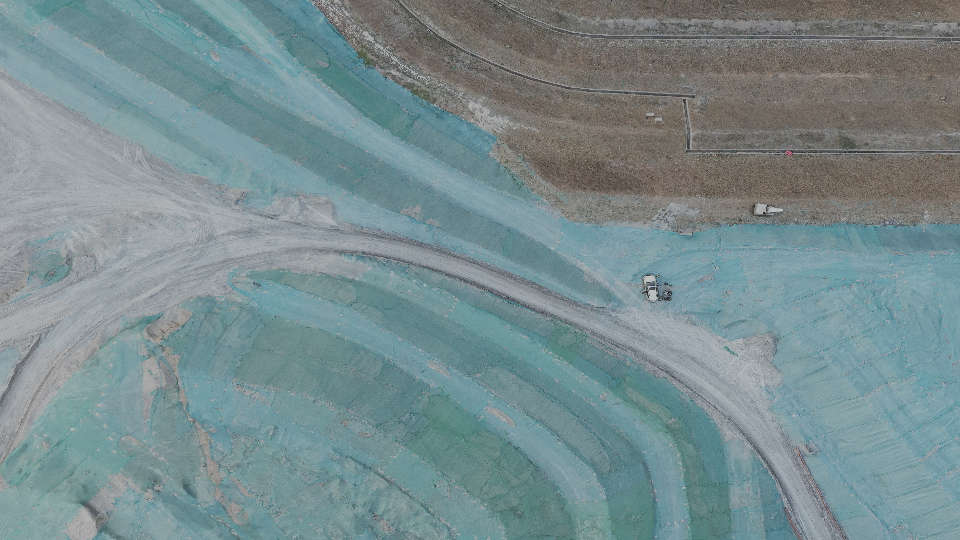}
			& 
\includegraphics[width=0.26\textwidth]{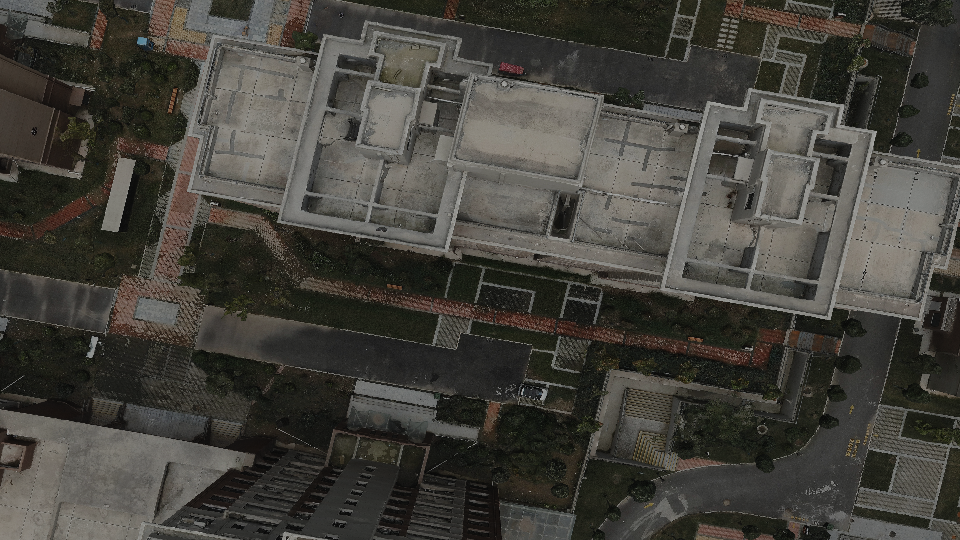}
		\\ 
    {\rotatebox{90}{Left disparity}}	&
\includegraphics[width=0.26\textwidth]{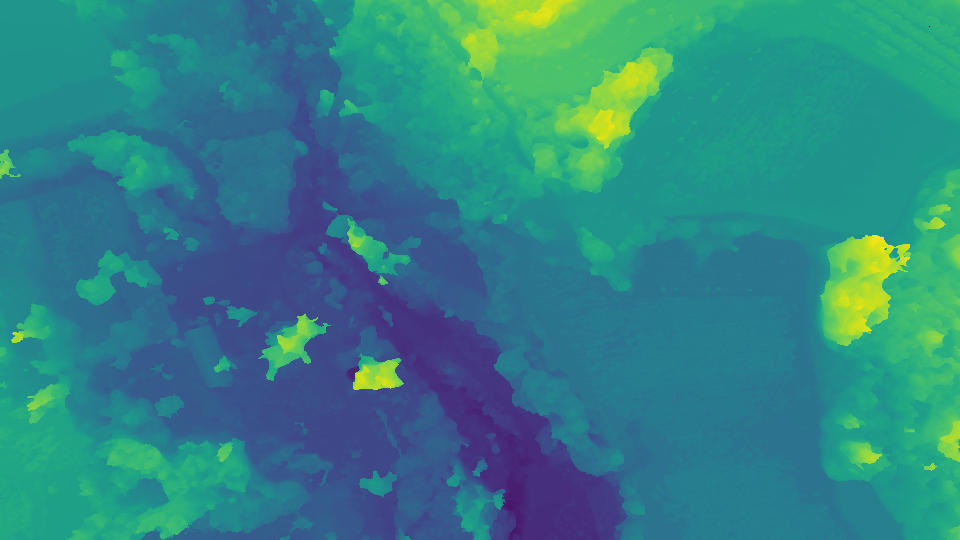}
			& 
\includegraphics[width=0.26\textwidth]{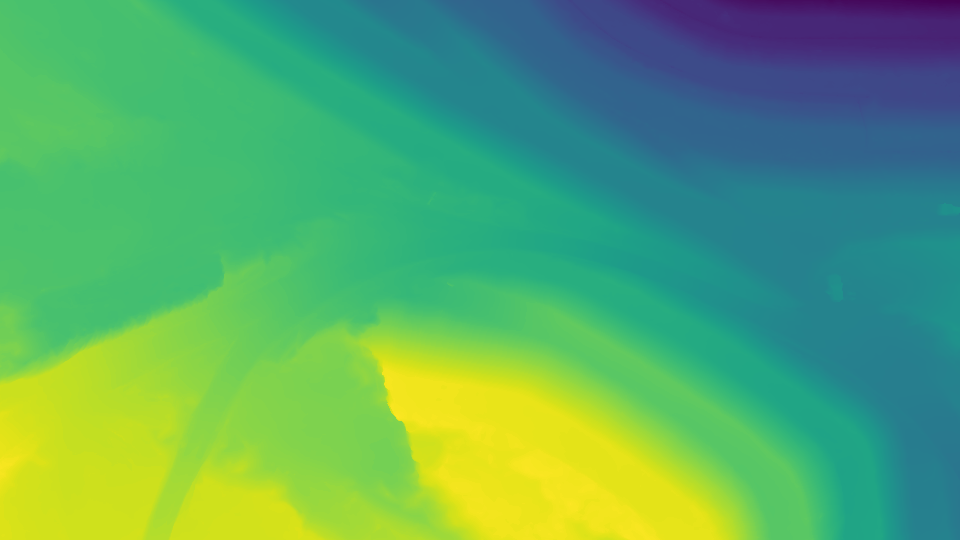}
			& 
\includegraphics[width=0.26\textwidth]{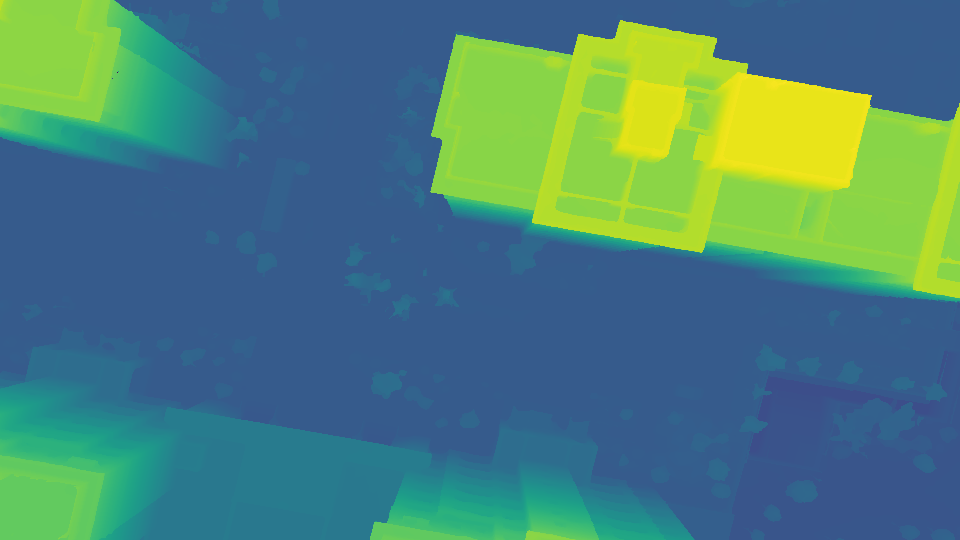}
		\\ 
    {\rotatebox{90}{Right disparity}}	&
\includegraphics[width=0.26\textwidth]{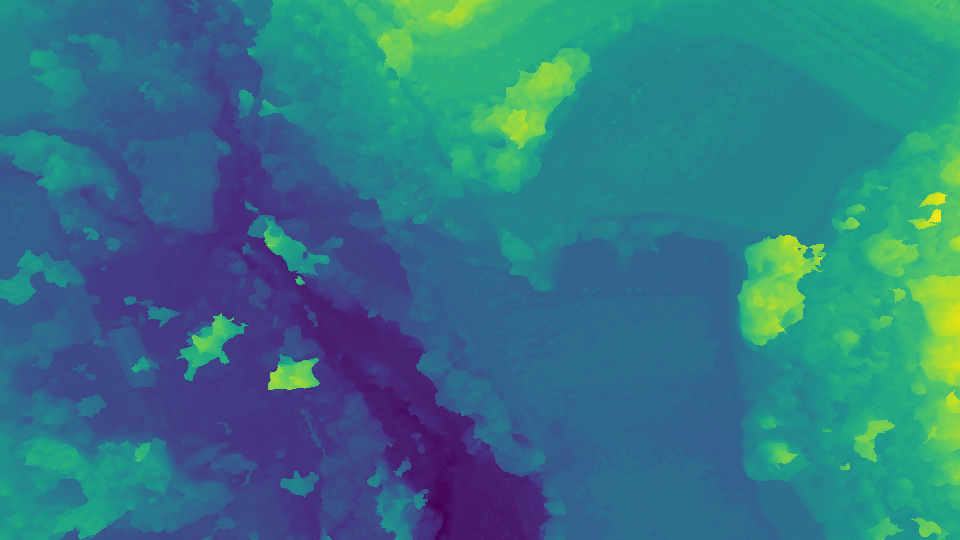}
			& 
\includegraphics[width=0.26\textwidth]{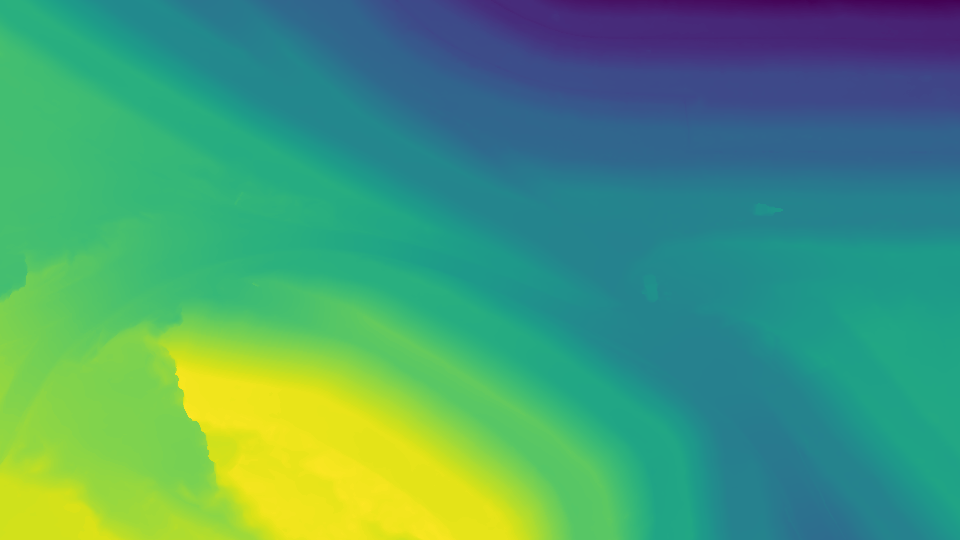}
			& 
\includegraphics[width=0.26\textwidth]{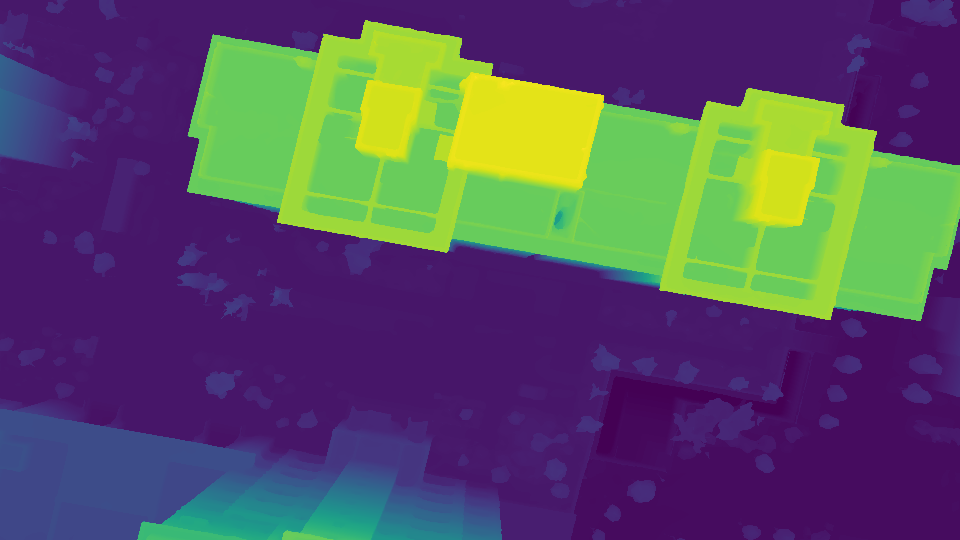}
\\ 
    {\rotatebox{90}{Disparity distribution}}	&
\includegraphics[width=0.26\textwidth]{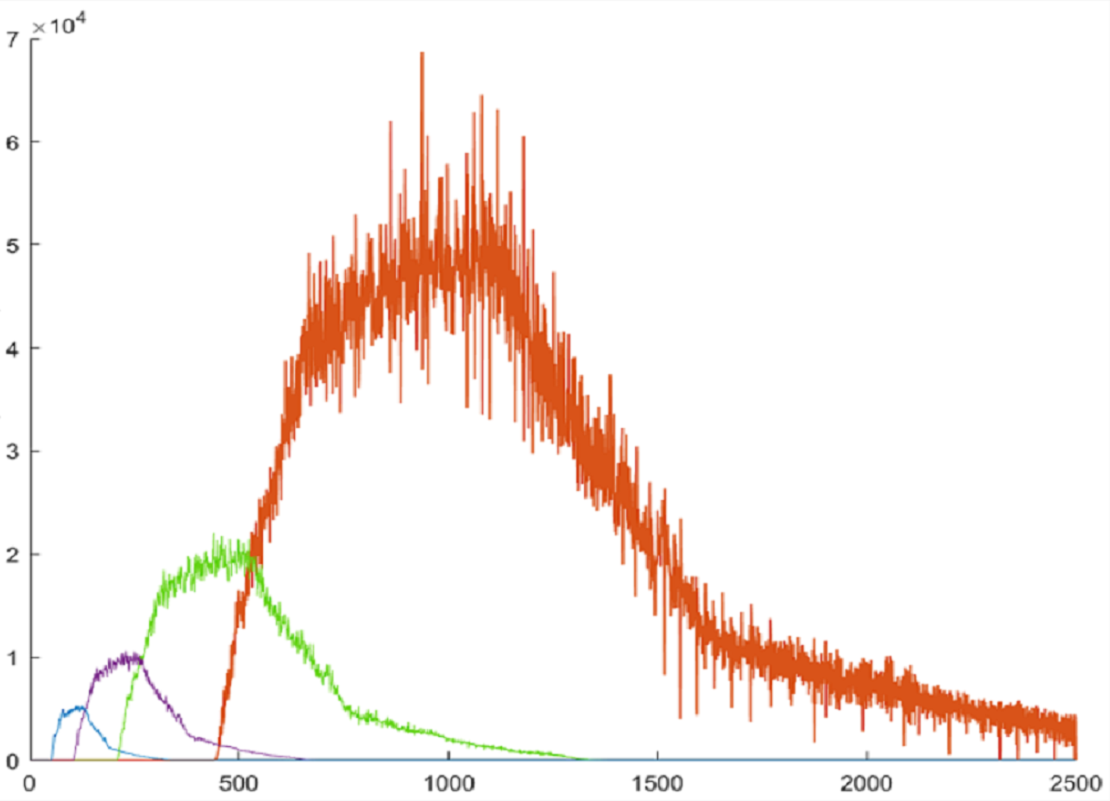}
			& 
\includegraphics[width=0.26\textwidth]{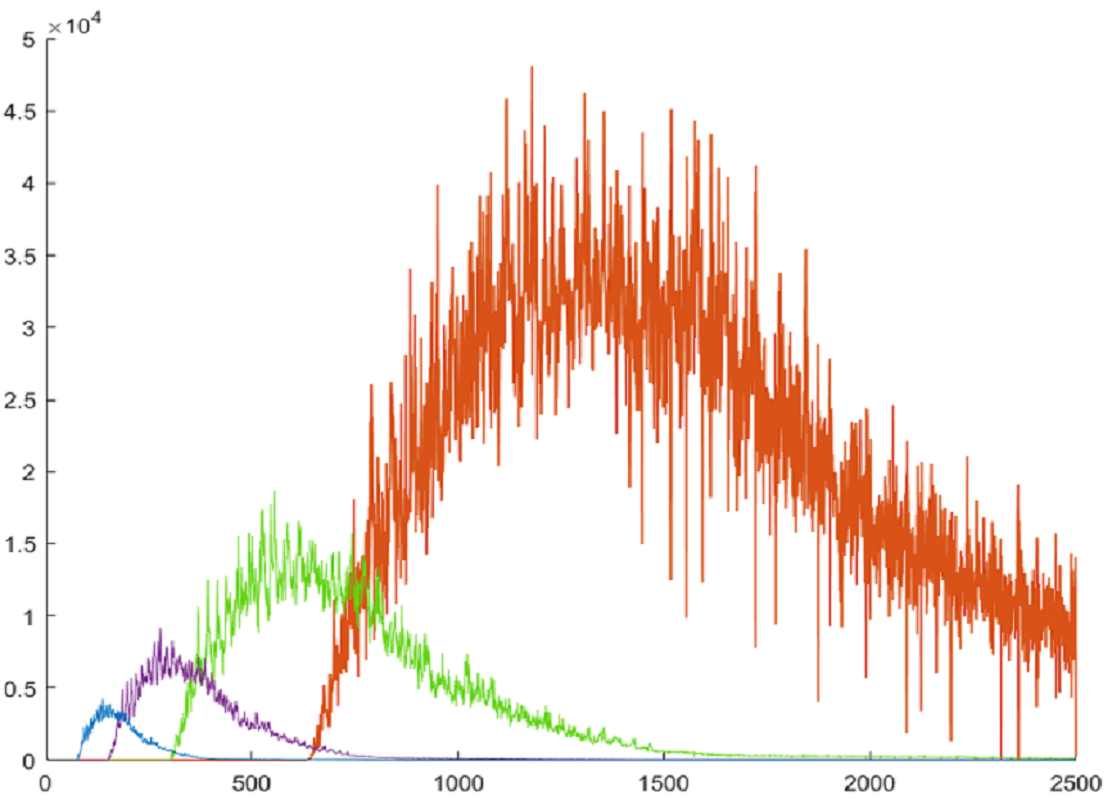}
	        & 
\includegraphics[width=0.26\textwidth]{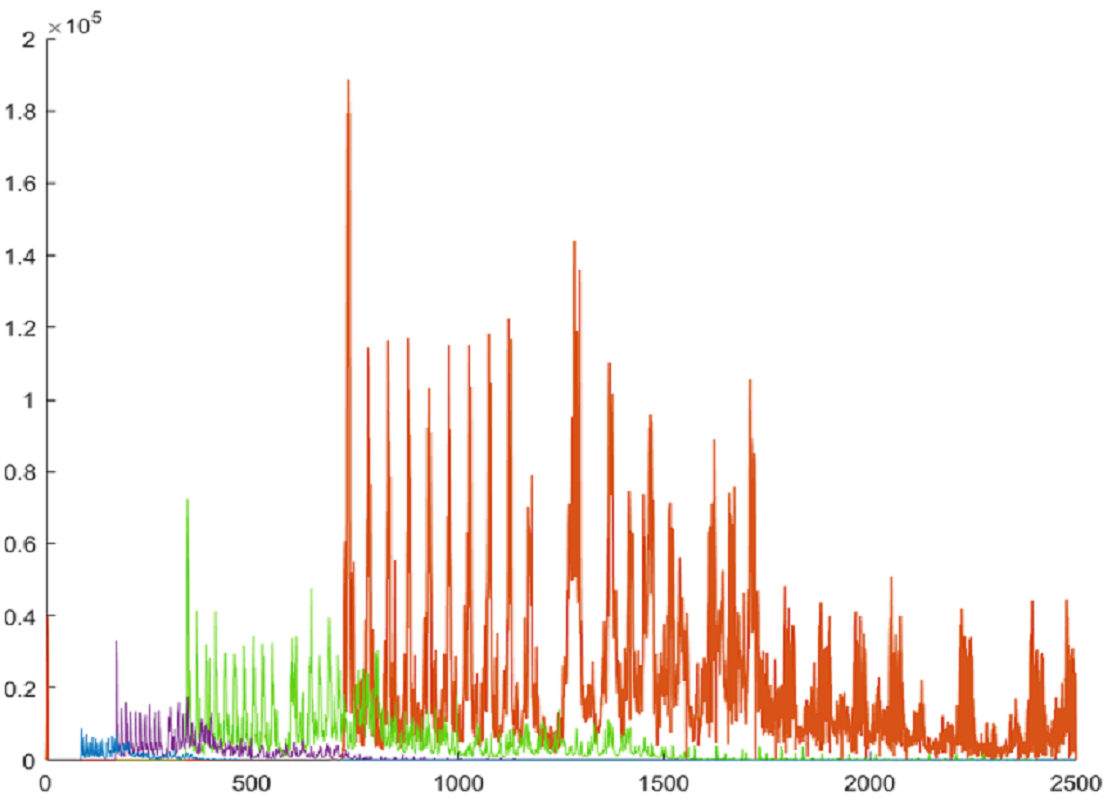}
\end{tabular} }  %

 \caption{UAVStereo Synthetic data. Left: Residential land. Center: Forest area. Right: Mining area. In distribution histogram, the blue, purple, green and orange line respectively represent the disparity distribution of 960 × 540 $px$, 1920 × 1080 $px$, 3840 × 2160 $px$, 8192 × 5460 $px$ disparity maps.}
\label{tab:samplesSynImages}
\end{table*}

\subsection{Real Dataset}
\label{real Dataset}
For real data, we make use of the images collected by P1, POS (position and orientation system) data containing image position and orientation, and the georeferenced OBJ models. The collected data generates epipolar stereo image pairs and corresponding disparity maps in a four-step pipeline: initial disparity maps generation, epipolar image generation, epipolar disparity maps generation, and post-processing. 

After aligning the camera with the model using the position $(Xs, Ys, Zs)$ and orientation $(\varphi, \omega, \kappa)$ in POS, an initial disparity map corresponding to the image can be rendered in the same way as the synthetic data using Blender.

The second step of the processing pipeline is to create epipolar image pairs from adjacent images with sufficient overlap. Stereo rectification can be implemented by feature extraction and matching, homography matrix calculation, and interpolation resampling, which Off-the-shelf functions in the OpenCV library can implement. The corresponding orientation parameters are generated to facilitate subsequent disparity map processing.

Then, in order to retain the same transformation between pictures and disparity maps, we deal with disparity maps by applying the orientation parameters from the previous step.

The well-chosen photos and the corresponding disparity maps are then croped to 960 × 540 $px$, which can be applicable to most networks.

The result image pairs and coeresponding disparity maps are shown in Tab. \ref{tab:samplesrealImages}

\begin{table*}[h]
\small

   \centering
		\newcommand{\tabincell}[2]{\begin{tabular}{@{}#1@{}}#2\end{tabular}}
		\centering
		\resizebox{\textwidth}{!}{
		\begin{tabular}{m{1.0cm}<{\centering}m{4.5cm}<{\centering}m{4.5cm}<{\centering}m{4.5cm}}

    {\rotatebox{90}{Left image}}	&
\includegraphics[width=0.26\textwidth]{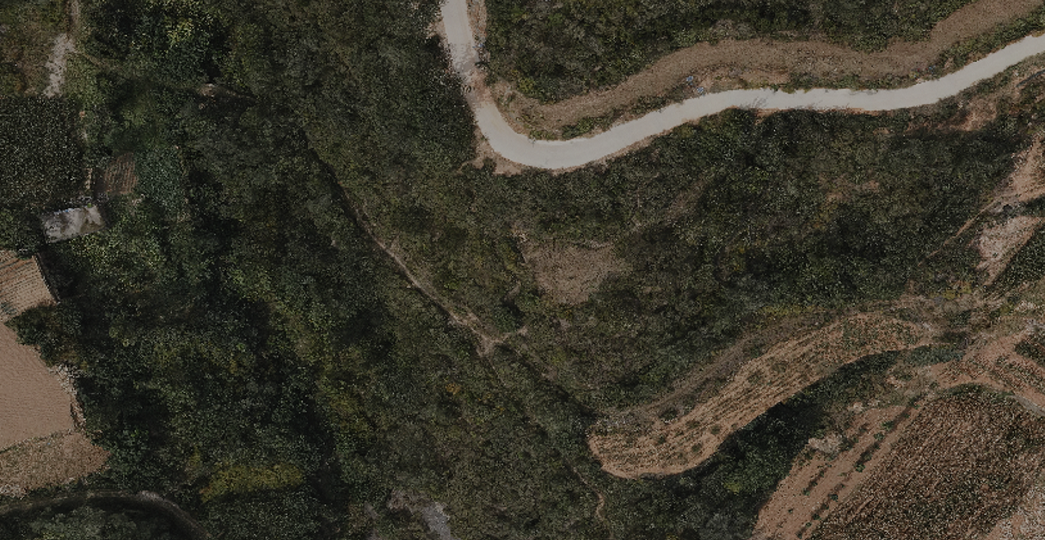}
			& 
\includegraphics[width=0.26\textwidth]{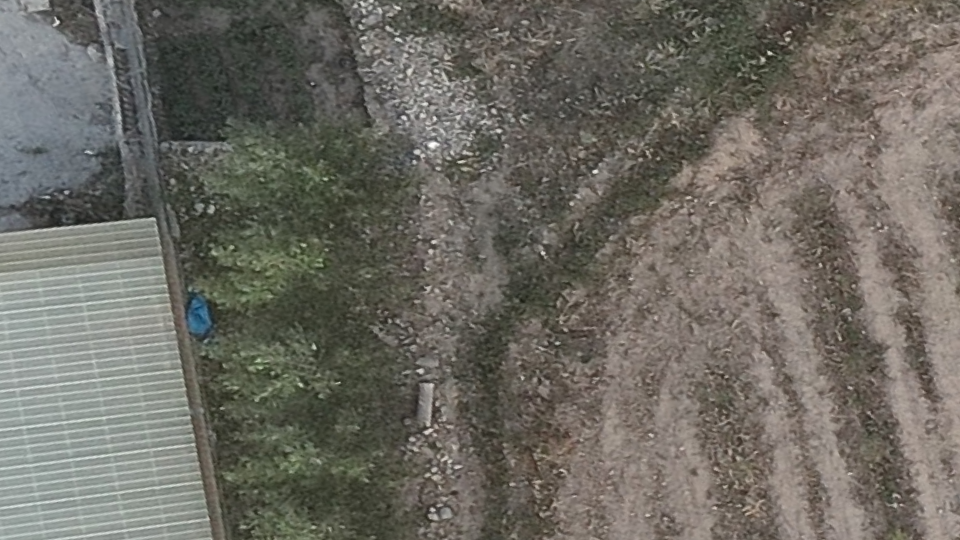}
	        & 
\includegraphics[width=0.26\textwidth]{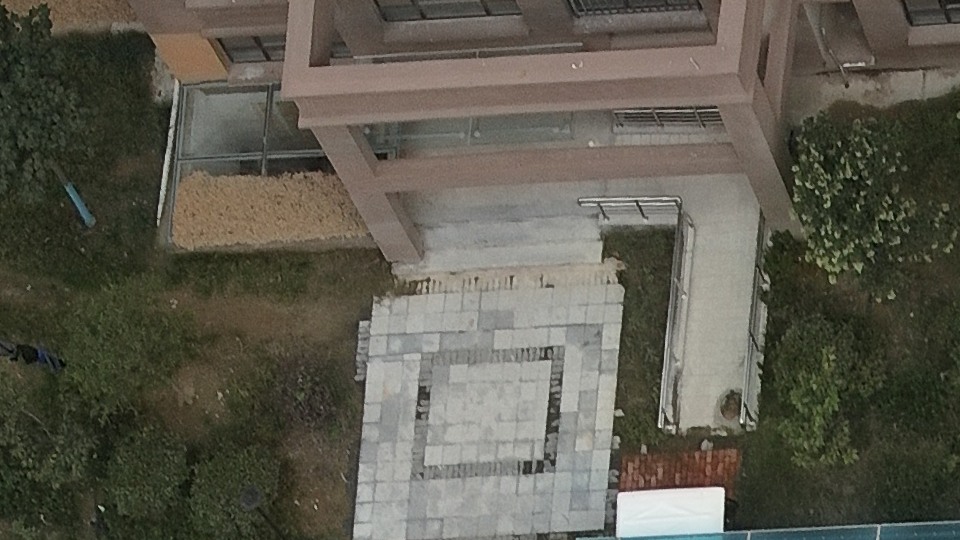}
		\\ 
	{\rotatebox{90}{Right image}}	&
\includegraphics[width=0.26\textwidth]{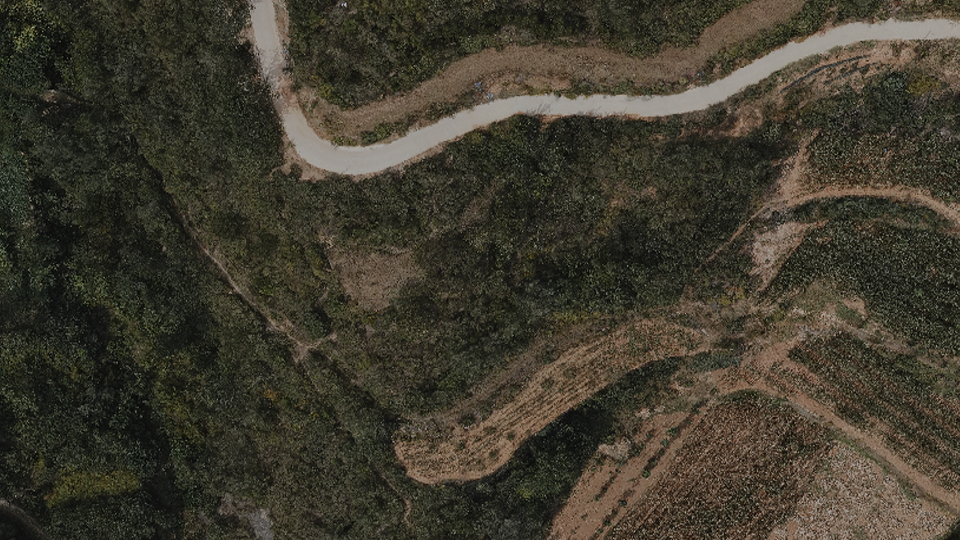}
			& 
\includegraphics[width=0.26\textwidth]{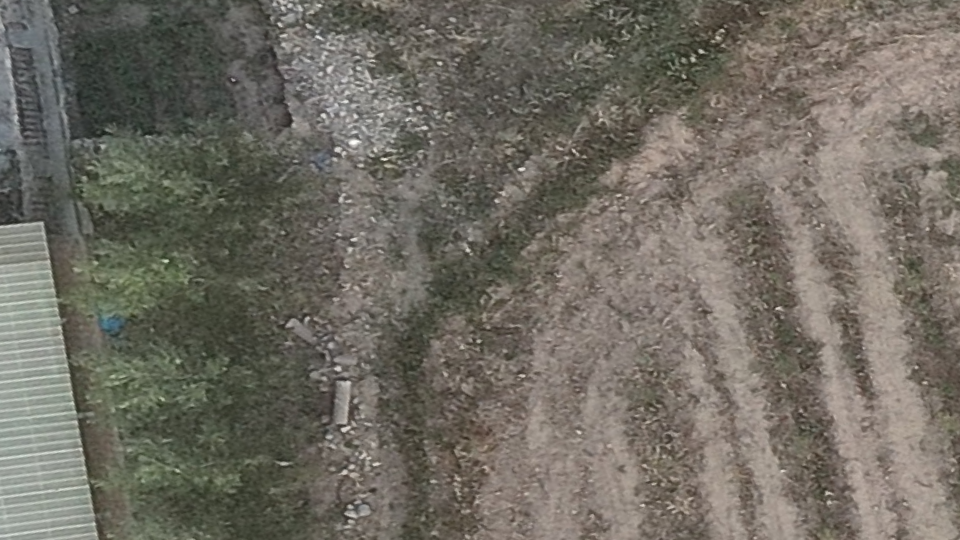}	
            &
\includegraphics[width=0.26\textwidth]{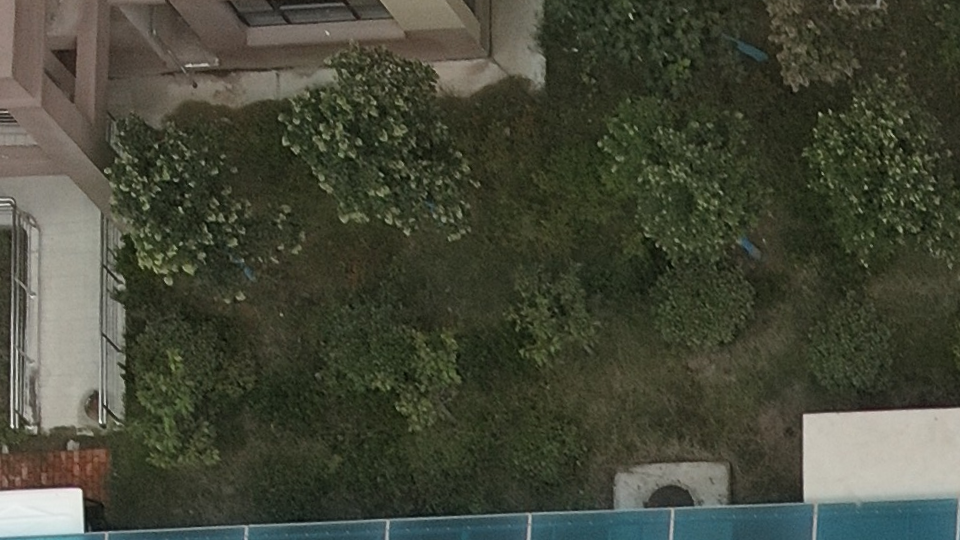}
		\\ 
    {\rotatebox{90}{Left disparity map}}	&
\includegraphics[width=0.26\textwidth]{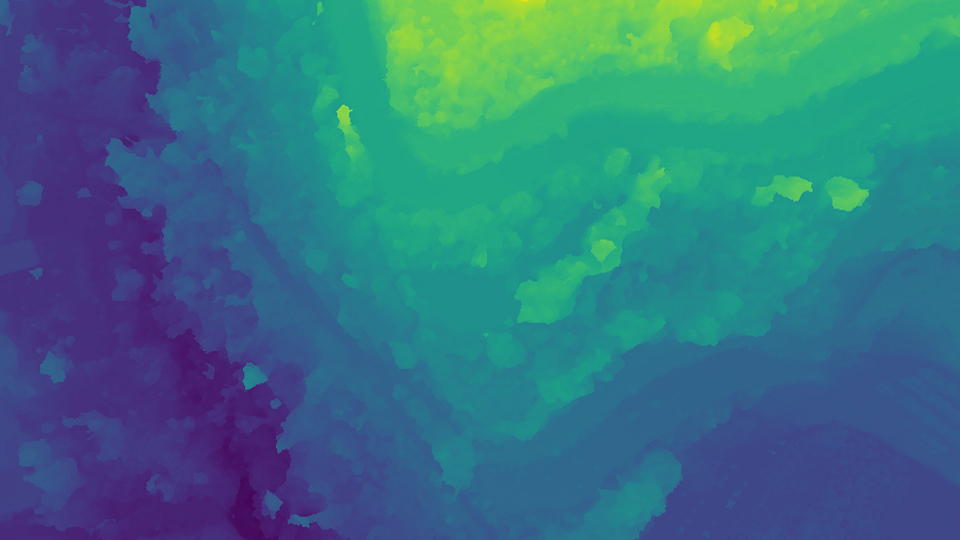}
	& 
\includegraphics[width=0.26\textwidth]{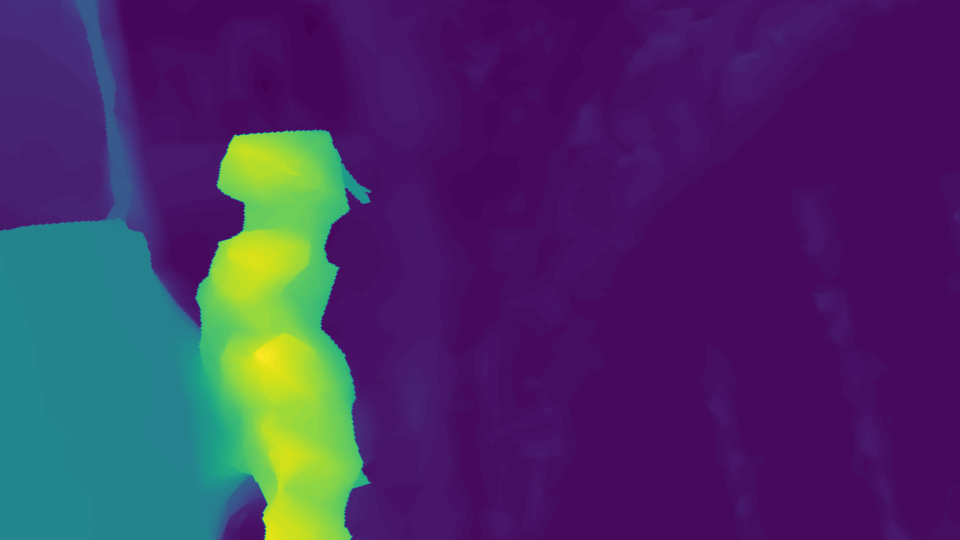}
	& 
\includegraphics[width=0.26\textwidth]{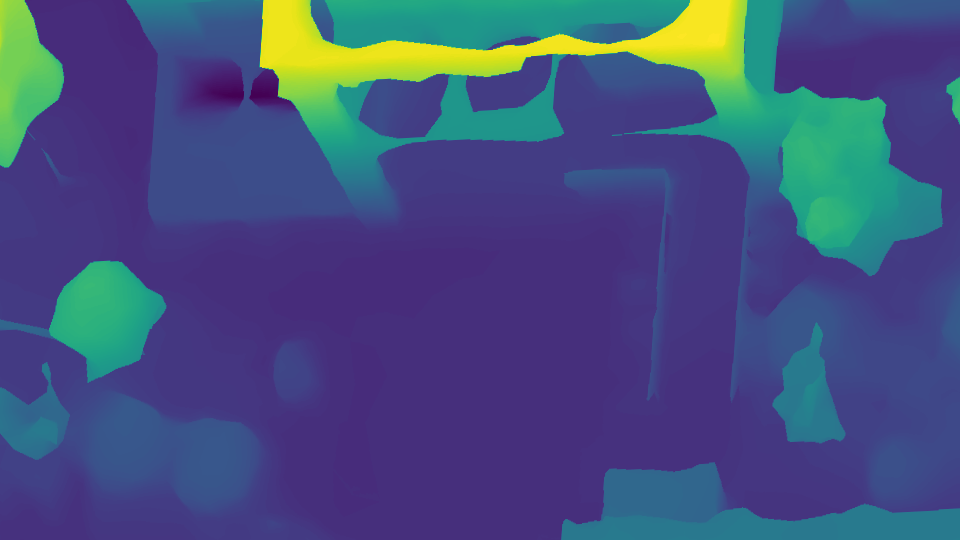}
		\\ 
    {\rotatebox{90}{Right disparity map}}	&
\includegraphics[width=0.26\textwidth]{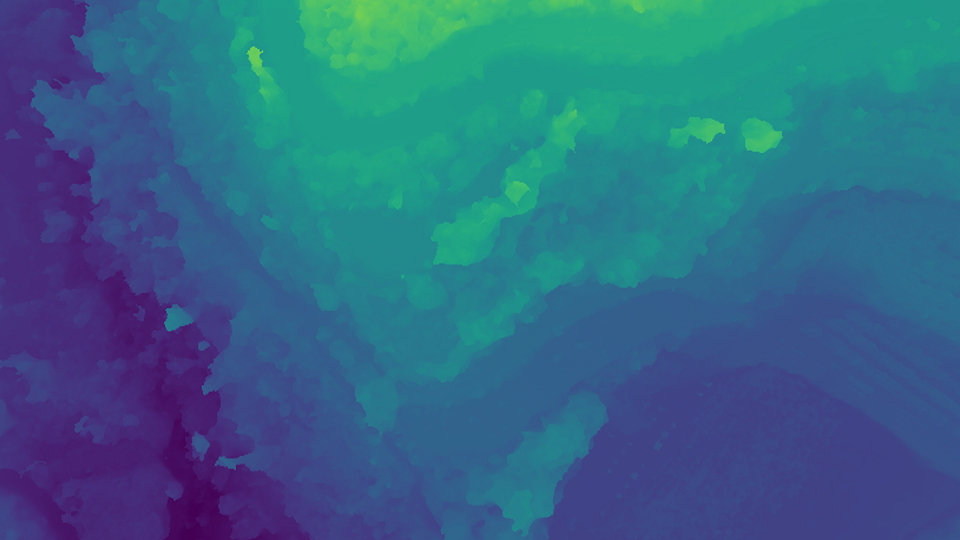}
	& 
\includegraphics[width=0.26\textwidth]{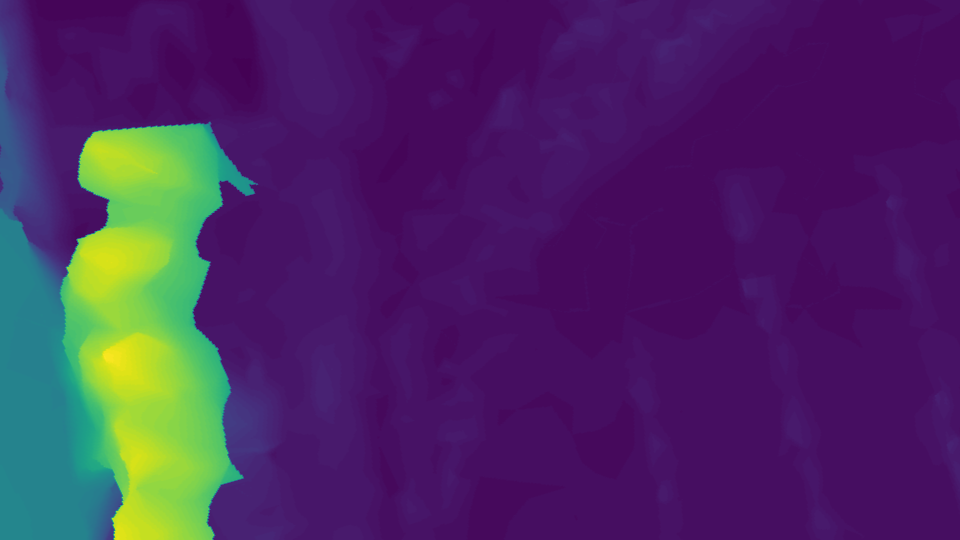}
	& 
\includegraphics[width=0.26\textwidth]{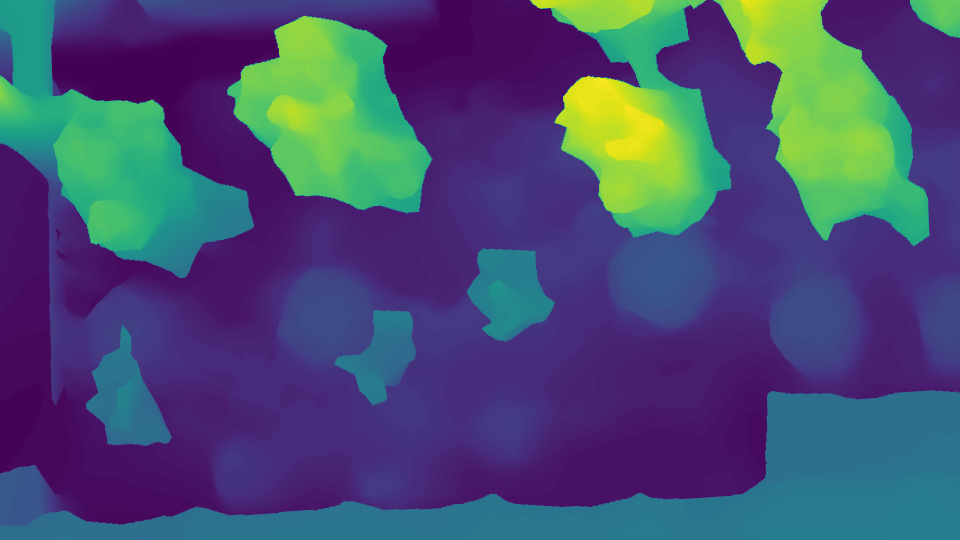}
\\   
    {\rotatebox{90}{Disparity distribution}}	&
\includegraphics[width=0.26\textwidth]{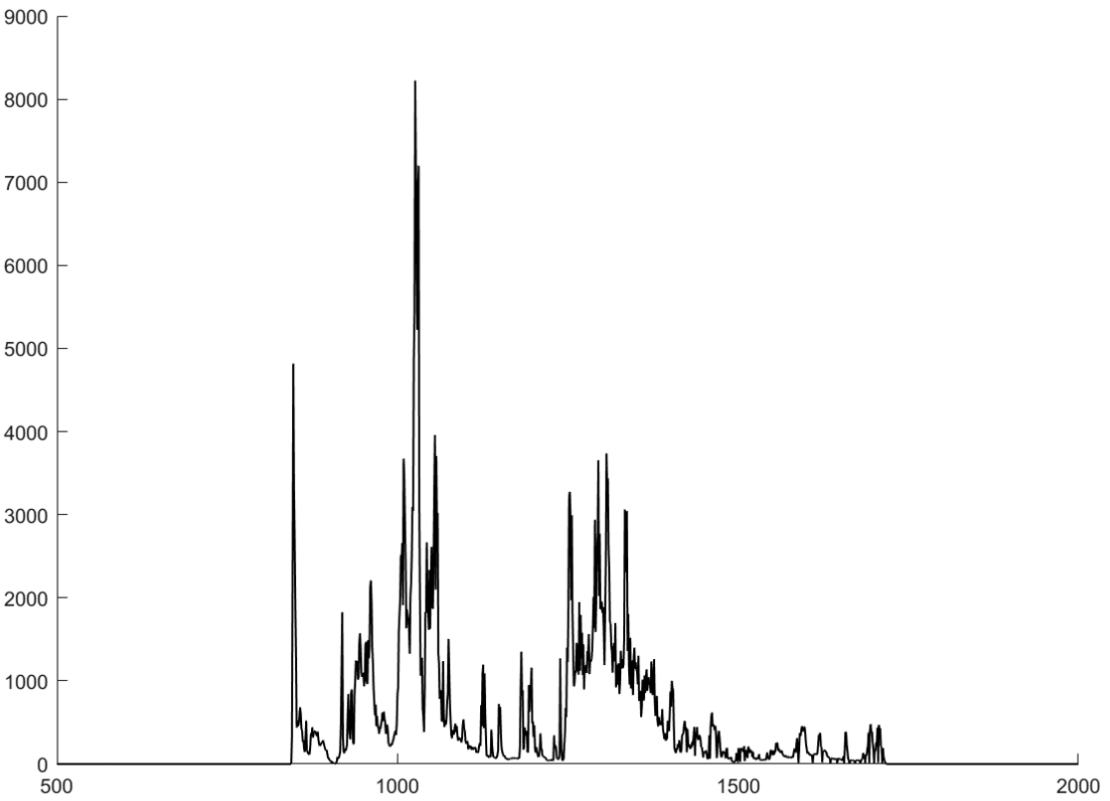}
			& 
\includegraphics[width=0.26\textwidth]{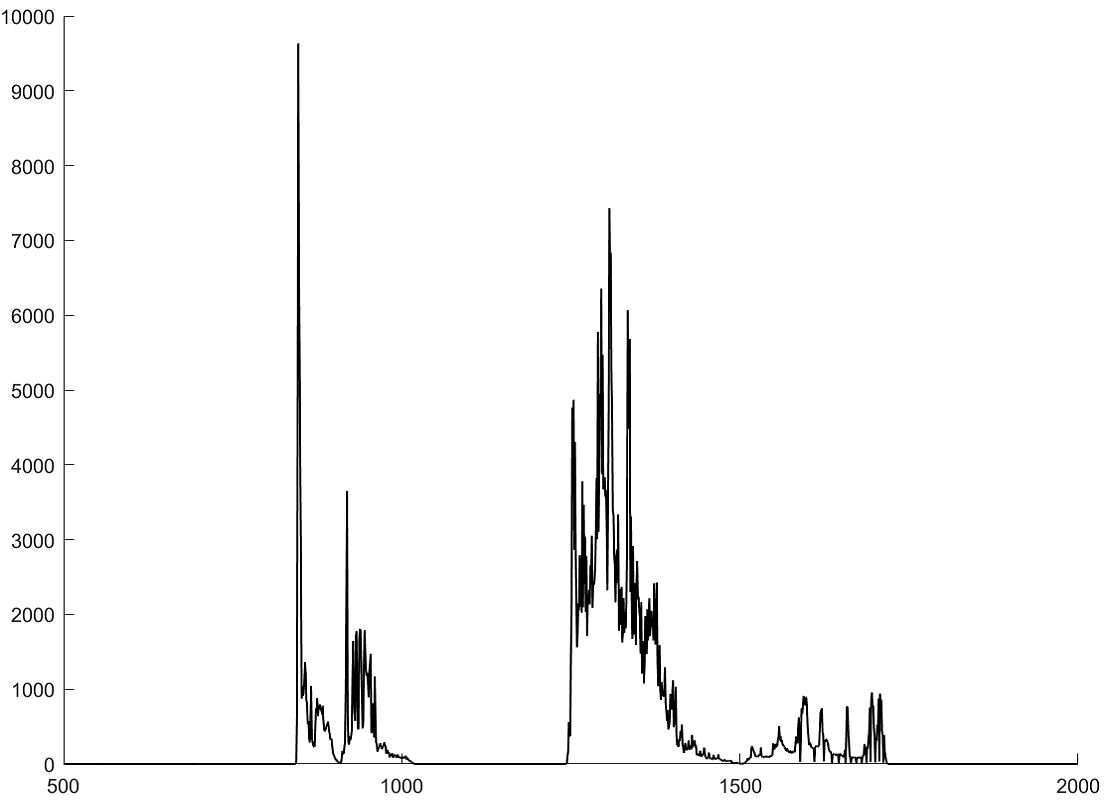}	
            &
\includegraphics[width=0.26\textwidth]{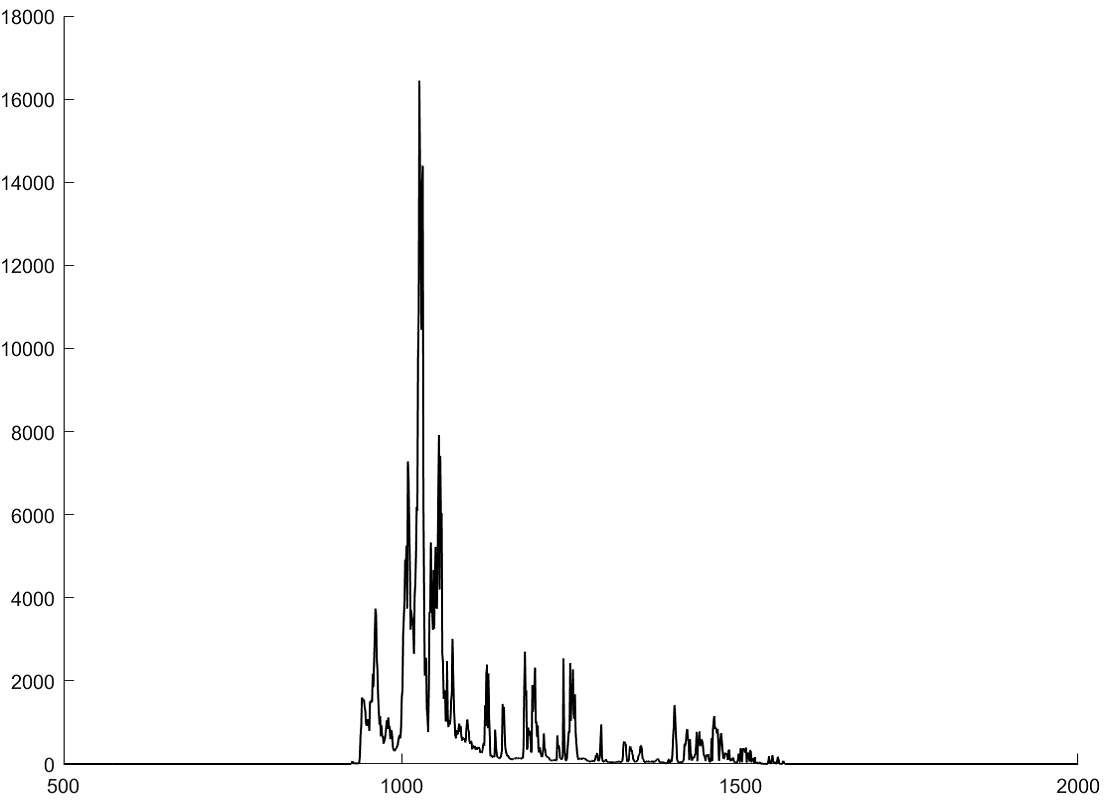}
\end{tabular} }  %

 \caption{UAVStereo Real data. Left: Residential land. Center: Forest area. Right: Mining area.}
\label{tab:samplesrealImages}
\end{table*}

\section{Experiences}
\label{sec:exp}

\subsection{Dataset Overview}
\label{Dataset Construction}
The above process creates a sizable stereo matching dataset containing real and synthetic data using various scene photographs and LiDAR data collected by the UAV platform. We divided the training set and testing set in a ratio of approximately 8:2 for each scenario, spliting the total 38781 stereo samples into 31024 and 7757 for training and testing proposes, detailed in Tab. \ref{tab:datasetFeature}. In Tab. \ref{tab:datasetFeature}, we also list key parameters and additional details of UAVStereo dataset. 

The contributions of UAVStereo can be summarized as follow: 
\begin{itemize}
\item First-ever UAV scenario stereo dataset. Different from autonomous driving, aerial, and indoor datasets, we propose a noval pipeline to generate images and disparity maps using UAV imagery and LiDAR point clouds in UAV scenarios.
\item Large disparity range. For the changes in imaging sensor and exploring areas, our dataset contains multiple resolution images in representative scenes to adapt to the changes of payload sensors and environments.
\item Containing both synthetic and real data. Compared with the existing single synthetic or real dataset, UAVStereo contains both synthetic and real datasets to bridge the gap between real and synthetic domains. 
\item High diversity. Our dataset provides a variety of representative scenarios and multiple flight paths, making it possible to account for most situations from the straight-down perspective of a drone.
\end{itemize}

\begin{table*}[h]
\small
   \caption{Key parameters of stereo data in UAVStereo.}

   \centering
   \renewcommand\arraystretch{1.3}
		\centering
		\begin{tabular}{m{2.25cm}<{\centering}m{2.25cm}<{\centering}m{2.25cm}<{\centering}m{2.25cm}<{\centering}m{2.25cm}<{\centering}m{2.25cm}<{\centering}m{2.25cm}<{\centering}}
  \toprule
\textbf{Scene type} &
Synthetic forest
			& 
Synthetic mining
			& 
Synthetic residential land
            & 
Real forest
            & 
Real mining
            & 
Real residential land
\\ \hline
\textbf{Stereo number}&
13320
			& 
8640
			& 
9000
            & 
1625
            & 
3147
            & 
3049
		\\ \hline
\textbf{Dataset split
(Training / Testing)}&
10656 / 2664 
			& 
6912 / 1728 
			& 
7200 / 1800 
            & 
1300 / 325 
            & 
2517 / 630 
            & 
2439 / 610 
		\\ \hline
\textbf{Baseline range}&
15 - 35 $m$
			& 
15 - 35 $m$
			& 
15 - 35 $m$
            & 
22.8 - 35.6 $m$
            & 
26.7 - 33.7 $m$
            & 
14.6 - 19.2 $m$
		\\ \hline
\textbf{Resolution}&
960 × 540 $px$
1920 × 1080 $px$
3840 × 2160 $px$
8192 × 5460 $px$
			& 
960 × 540 $px$
1920 × 1080 $px$
3840 × 2160 $px$
8192 × 5460 $px$
			& 
960 × 540 $px$
1920 × 1080 $px$
3840 × 2160 $px$
8192 × 5460 $px$
            & 
960 × 540 $px$
            & 
960 × 540 $px$
            & 
960 × 540 $px$
\\    
\bottomrule
\end{tabular}  %
\label{tab:datasetFeature}
\end{table*}

\subsection{Experience Setup}
\label{Experience Setup}
After generating the epipolar images and the corresponding ground truth disparity maps, we evaluate several traditional and learning-based methods on our UAVStereo dataset. Comparative results illustrate the significance and challenge of our dataset.

In our experiments, we run a set of deep learning methods on UAVStereo in order to assess their accuracy, including PSMNet \cite{chang2018pyramid}, DSMNet \cite{zhang2020domain}, CFNet \cite{shen2021cfnet}, RAFT-Stereo \cite{lipson2021raft}, EAIStereo \cite{Zhao_2022_ACCV}. As references, we also evaluate the popular Semi-Global Matching algorithm (SGM) \cite{hirschmuller2007stereo} in its fast variant.

We implemented the above deep neural network in PyTorch. All networks are trained end-to-end, given the images as input and the disparity maps. We used the same learning rate, optimizer and loss function as the original network. Since deep networks inferences are performed on a NVIDIA 3090 RTX GPU, we set the batch size to 1 and crop the image to 256×512 $px$ in all network training phase. The training process continues until the loss function is no longer changing. The last epoch model was used for evaluation.

The traditional SGM is available in OpenCV and easy to implement with C++. In SGM, we used census to calculate the matching cost, aggregated the matching cost on 8 paths. Post-processing such as consistency check, uniqueness constraint and culling of small connected regions were adopted to the completeness and consistency of output. 

To assess the accuracy of stereo algorithms and networks, we conducted quantitative statistics EPE (End Point Error) and NPE (N-pixel Error) as evaluation metrics. EPE is the absolute mean of the difference between the estimated disparity map and the ground truth; NPE is the percentage of pixels having error larger than a threshold N:
$$\operatorname{EPE}=\frac{1}{m} \sum_{i=1}^{m}\left|D_{pred}-D_{gt}\right|$$
$$\operatorname{NPE}=\frac{count(\left|D_{pred}-D_{gt}\right|>N)}{m}$$
$m$ is the total number of pixels, $D_{pred}$ is the output predicted disparity map, $D_{gt}$ is the ground truth disparity map. As initially our ground-truth disparity maps are inferred at 960 × 540 $px$, we assumed 3 pixels as the lowest threshold. Then, given the much larger disparity in real subset, we computed error rates up to 30PE and 100PE.

\subsection{Evaluation on Synthetic subset}
\label{sec:syn test}

In the evaluation of the synthetic subset, we trained the network on all training data instead of a single dataset to avoid overfitting on a particular subset. The pre-trained models was verified on different scenarios. Tab. \ref{tab:Syntest} compares the predicted disparity maps with its corresponding disparity ground-truth, collecting the outcome of this evaluation. 
In the evaluation of low-resolution(the upper portion of Tab. \ref{tab:Syntest}), the experimental results indicate that the traditional SGM method has considerable errors with UAV image pairs, whereas end-to-end networks significantly improve the stereo matching accuracy. 
Applying the SGM algorithm to UAV images, the output disparity maps are incomplete and discontinuous, resulting in large error metrics. Comparing the performance of SGM on Middlebury, we demonstrate that the SGM algorithm is unsuitable for processing UAV geographic images containing ill-areas, because this method actually uses a matching window of limited size, which is incapable of obtaining and utilizing global information.
Among leaning-based methods, PSMNet achieves the best results with EPE and 3PE values of 3.443 $px$ and 11.634 $\%$, respectively. The result demonstrates that our UAVStereo dataset can be applied for stereo matching networks,  with precision result comparable to that of blendedMVS \cite{yao2020blendedmvs} when converted into depth.
Despite the fact that the stereo matching network has grown rapidly since 2018, we found that PSMNet performed best on low-resolution images for challenging data, such as low-altitude drone images, possibly due to the use of global information through spatial pyramid pooling and 3D convolution strategies. Meanwhile, we point out that the latest network EAIStereo network's loss function converges to the large loss value, leading to large errors on the testing set. Consequently, EAIStereo may not be suitable for UAV images.

\begin{table*}[!h]
\small
   \caption{\textbf{Results on the UAVStereo Synthetic subset.} We trained model on 960×540 $px$ and evaluated the model on multiple resolution ground-truth maps. Best scores in \textbf{bold}. R: Residential land testing subset, F: Forest areas testing subset, M: Mining areas testing subset, A: All synthetic testing set.}
   \label{tab:Syntest}

   \centering
   \renewcommand\arraystretch{1.3}
		\centering
		\begin{tabular}{cccccccccccc}
    \hline
    \multirow{2}{*}{\textbf{Method}} & \multirow{2}{*}{\textbf{Resolution} }   & \multicolumn{2}{c}{\textbf{R}}  & \multicolumn{2}{c}{\textbf{F}} & \multicolumn{2}{c}{\textbf{M}} & \multicolumn{2}{c}{\textbf{A}}
\\  \cline{3-10}
&  & \textbf{EPE ($px$)}&  \textbf{3PE ($\%$)}& \textbf{EPE ($px$)} & \textbf{3PE ($\%$)} & \textbf{EPE ($px$)}& \textbf{3PE ($\%$)}&\textbf{EPE ($px$)} &\textbf{3PE ($\%$)}  \\ \hline
SGM \cite{hirschmuller2007stereo} &960×540 $px$ &102.035	&92.577	&135.767 &96.125 &69.484 &92.756 &102.428	&93.819\\
PSMNet \cite{chang2018pyramid} &960×540 $px$	&\textbf{4.688}	&\textbf{15.701}	&\textbf{4.421}	&\textbf{15.057}	&\textbf{4.084}	&\textbf{15.031}	&\textbf{3.443}	&\textbf{11.634}	\\
DSMNet \cite{zhang2020domain} &960×540 $px$	&9.003	&53.434	&6.861	&44.306	&5.317	&35.918	&4.482	&24.423	\\
CFNet \cite{shen2021cfnet} &960×540 $px$	&14.995	&48.621	&5.509	&39.131	&10.019	&36.383	&7.371	&42.554	\\
RAFT-Stereo \cite{lipson2021raft} &960×540 $px$	&15.405	&18.127	&6.769	&19.189	&15.685	&21.464	&3.924	&12.295	\\
EAIStereo \cite{Zhao_2022_ACCV} &960×540 $px$	&56.698	&95.732	&110.512	&98.943	&152.666	&99.254	&110.512	&98.988	\\
 \hline \hline
PSMNet \cite{chang2018pyramid} &1920×1080 $px$	&\textbf{12.495}	&28.090	&13.280	&\textbf{23.177}	&14.346	&29.795	&10.274	&21.558	\\
DSMNet \cite{zhang2020domain} &1920×1080 $px$	&21.657	&49.312	&\textbf{7.295}	&55.982	&38.843	&43.053	&16.800	&29.772	\\
CFNet \cite{shen2021cfnet} &1920×1080 $px$	&177.395	&68.826	&15.359	&58.092	&105.355	&69.150	&68.445	&76.356	\\
RAFT-Stereo \cite{lipson2021raft} &1920×1080 $px$	&30.806	&\textbf{27.113}	&13.530	&28.766	&\textbf{31.368}	&\textbf{27.784}	&\textbf{7.848}	&\textbf{18.311}	\\
 \hline \hline
PSMNet \cite{chang2018pyramid} &3840×2160 $px$	&\textbf{28.485}	&52.242	&38.918	&\textbf{26.221}	&\textbf{24.691}	&48.805	&20.210	&39.473	\\
DSMNet \cite{zhang2020domain} &3840×2160 $px$	&36.550	&59.702	&\textbf{10.581}	&62.935	&69.968	&45.078	&29.923	&38.377	\\
CFNet \cite{shen2021cfnet} &3840×2160 $px$	&217.121	&69.205	&30.718	&73.103	&179.721	&78.928	&137.193	&87.522	\\
RAFT-Stereo \cite{lipson2021raft} &3840×2160 $px$	&61.613	&\textbf{42.941}	&27.062	&44.158	&62.736	&\textbf{40.351}	&\textbf{15.697}	&\textbf{29.806}	\\
 \hline 
\end{tabular}  
\end{table*}

We also list representive disparity maps predicted by above algorithms in \ref{tab:result}. It can be seen that the disparity maps generated by SGM exist quite a few invalid values, resulting in large error metrics. While learning-based models can inference complete and continuous disparity maps in challenging regions like textureless ground. This result shows the capability superiority of deep learning-based stereo matching on UAV images. Among the learning-based algorithms, PSMNet and RAFT-Stereo perform better in the dataset, since accurate disparity maps can be obtained in all three scenarios.

\begin{table}[!htp]
\small
   \caption{\textbf{Results on the UAVStereo Real subset.}}
   \label{tab:Realtest}

   \centering
   \renewcommand\arraystretch{1.3}
		\centering
		\begin{tabular}{cccc}
    \hline
&  \textbf{EPE ($px$)}&  \textbf{30PE ($\%$)}& \textbf{100PE ($\%$)} \\ \hline
Synthetic &175.752	&88.717 &75.538	\\
Real &111.236	&71.867	&48.288	\\
Finetuned &101.386	&72.023 & 	52.203\\
 \hline 
\end{tabular}  
\end{table}

\begin{table*}[!h]
\small

   \centering
		\newcommand{\tabincell}[2]{\begin{tabular}{@{}#1@{}}#2\end{tabular}}
		\centering
		\resizebox{\textwidth}{!}{
		\begin{tabular}{m{4.5cm}<{\centering}m{4.5cm}<{\centering}m{4.5cm}<{\centering}}

\includegraphics[width=0.26\textwidth]{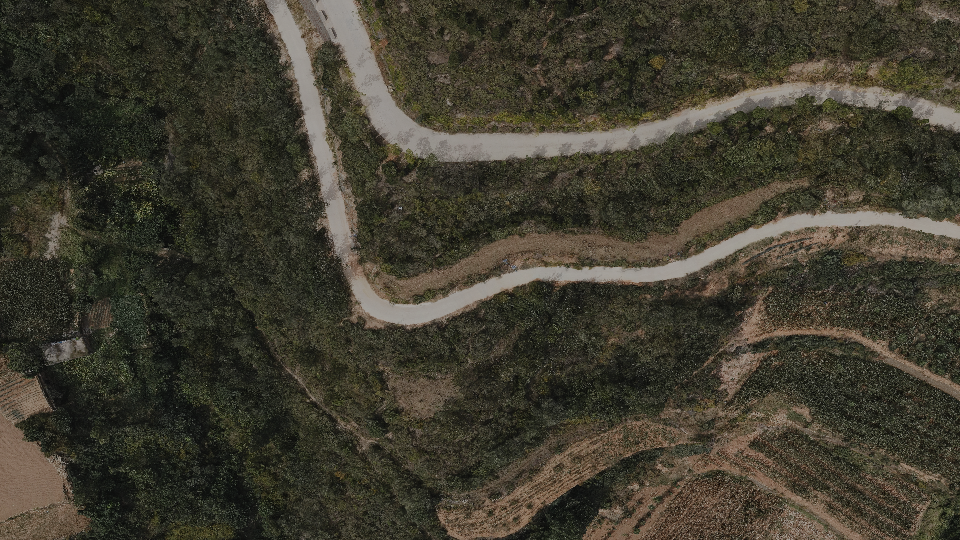}
			& 
\includegraphics[width=0.26\textwidth]{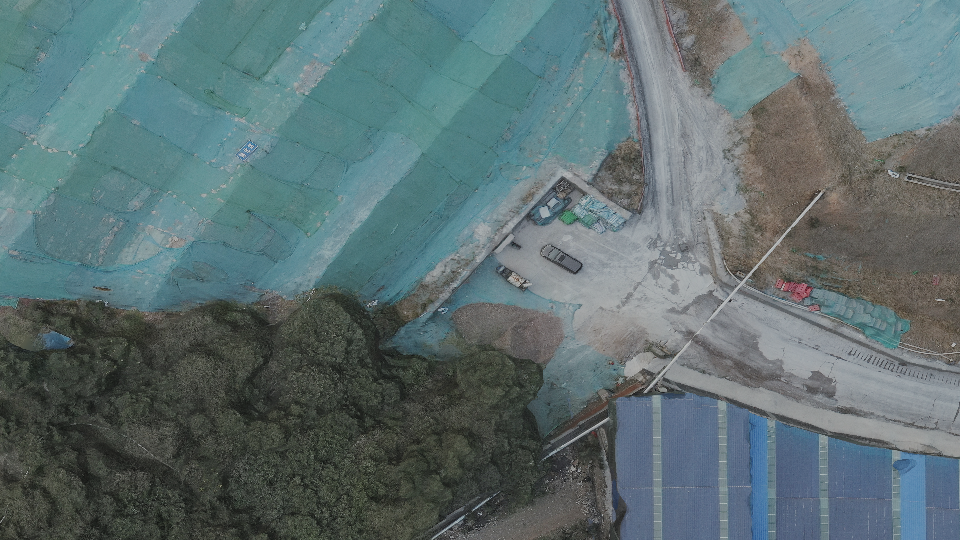}
	        & 
\includegraphics[width=0.26\textwidth]{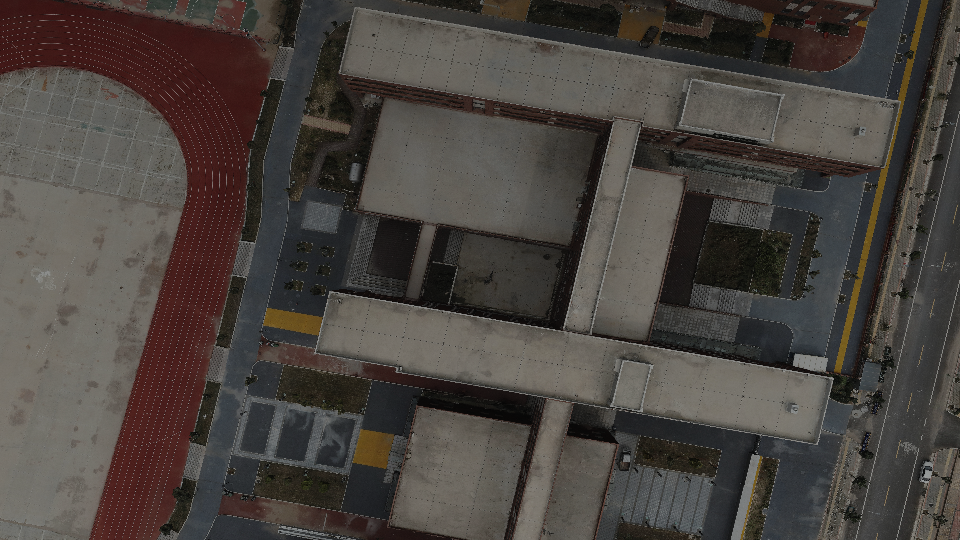}
		\\ 

\includegraphics[width=0.26\textwidth]{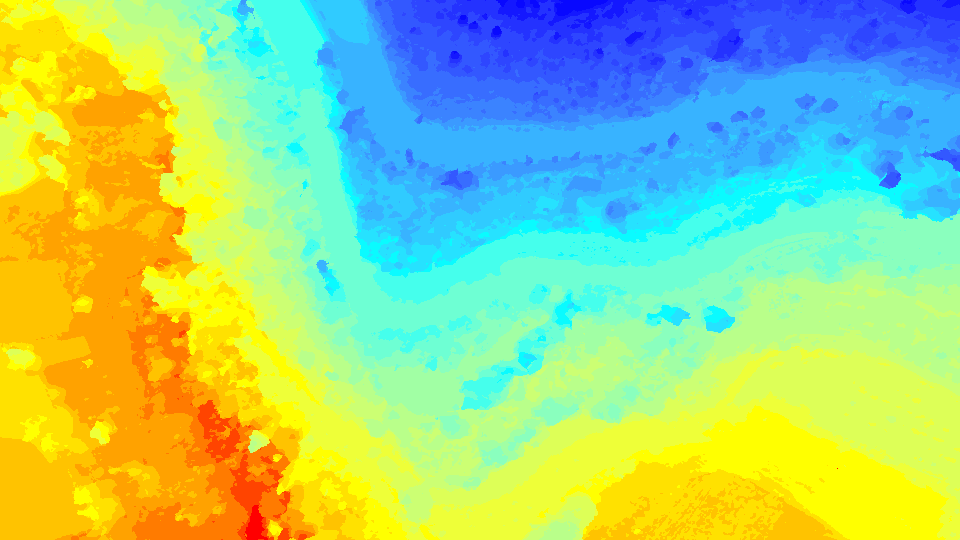}
			& 
\includegraphics[width=0.26\textwidth]{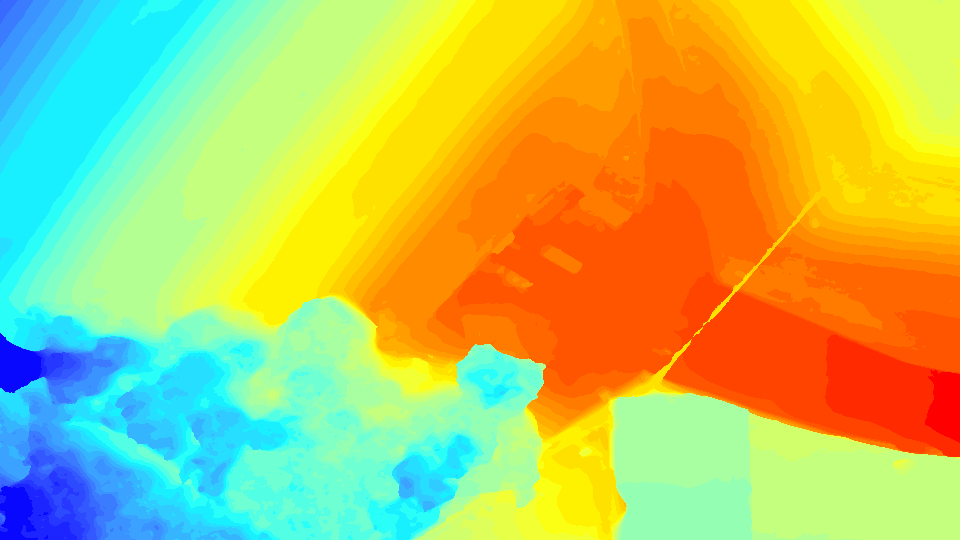}	
            &
\includegraphics[width=0.26\textwidth]{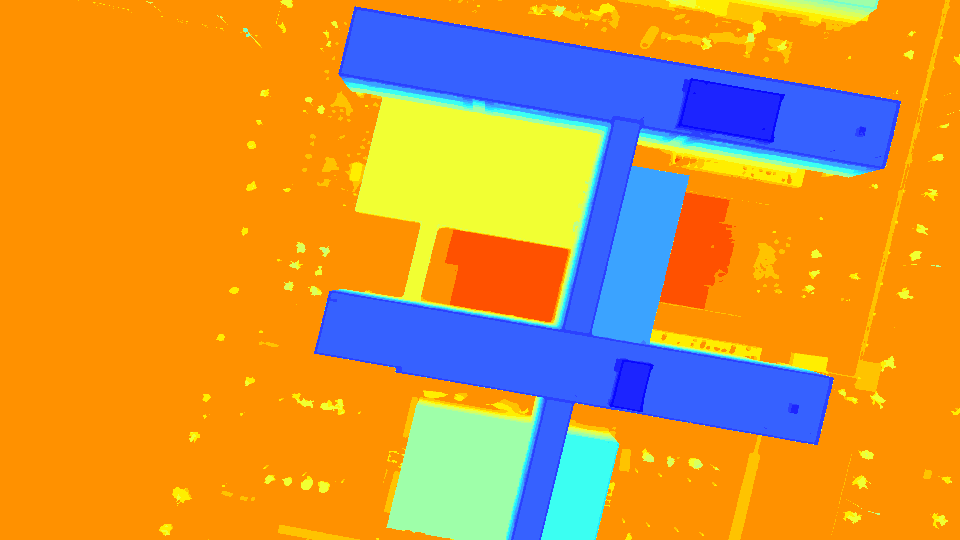}
		\\ 

\includegraphics[width=0.26\textwidth]{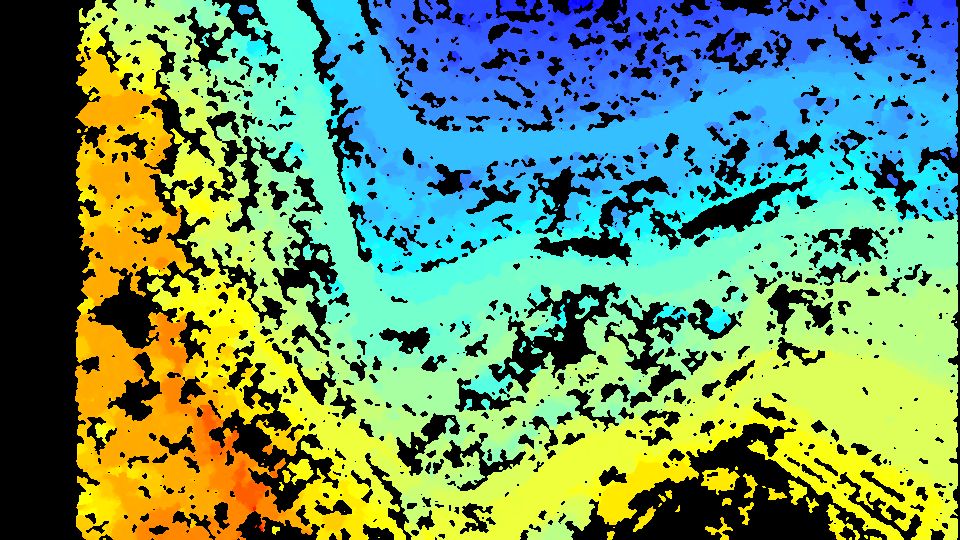}
	& 
\includegraphics[width=0.26\textwidth]{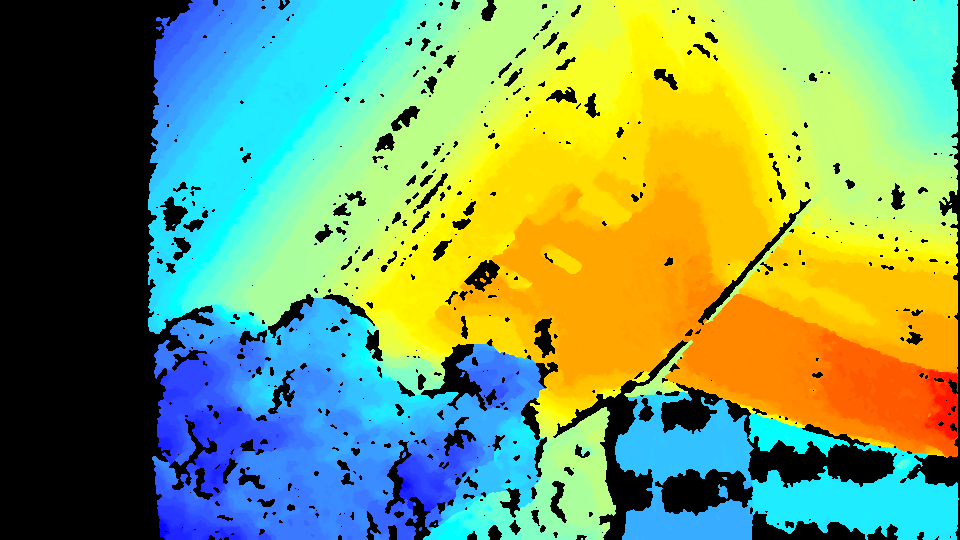}
	& 
\includegraphics[width=0.26\textwidth]{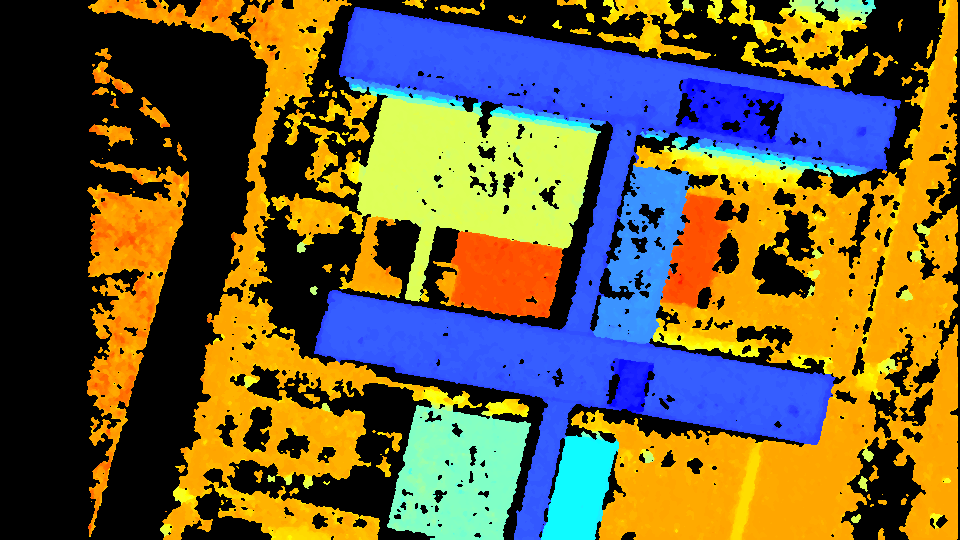}
		\\ 

\includegraphics[width=0.26\textwidth]{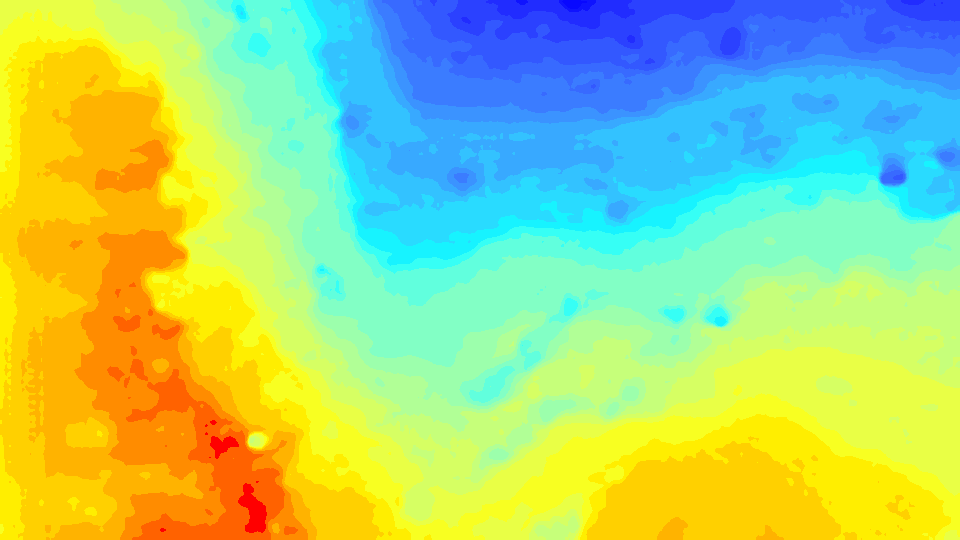}
	& 
\includegraphics[width=0.26\textwidth]{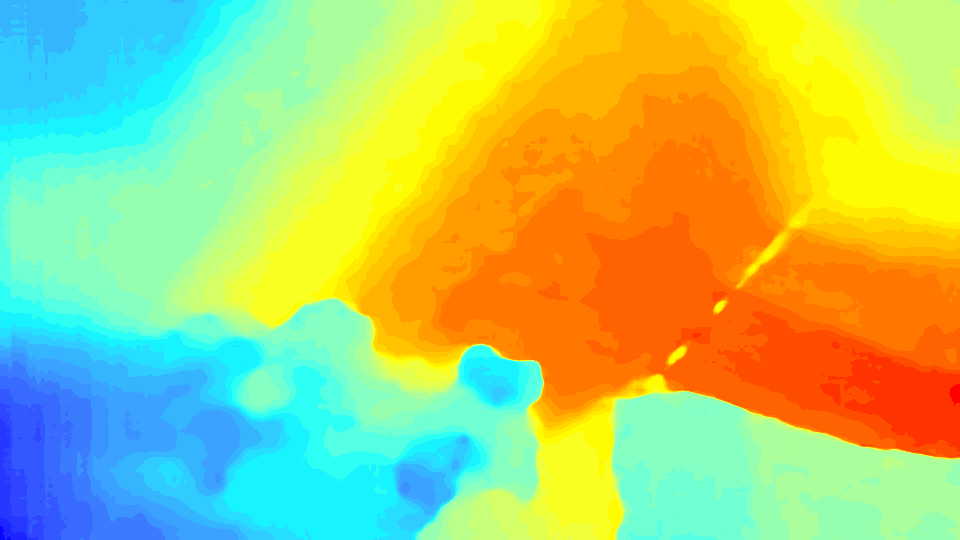}
	& 
\includegraphics[width=0.26\textwidth]{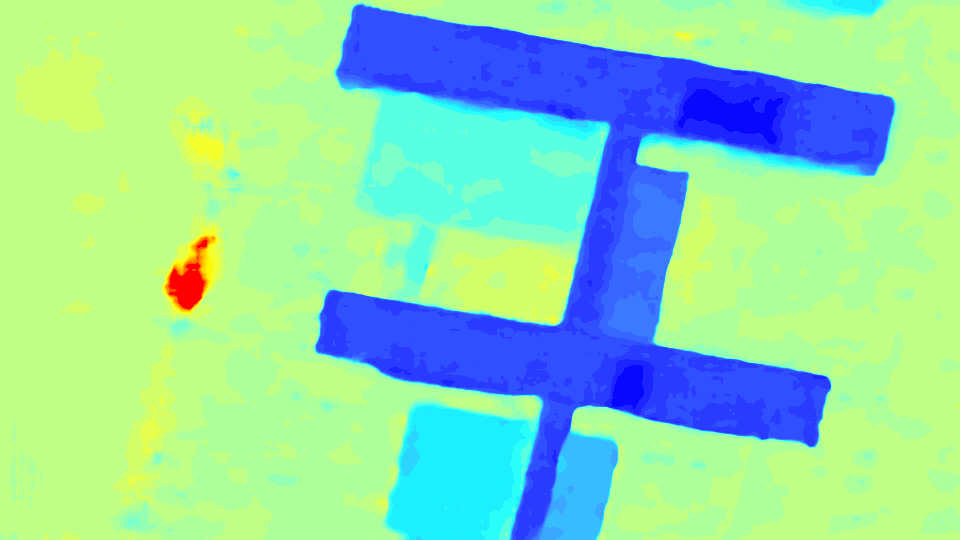}
\\   

\includegraphics[width=0.26\textwidth]{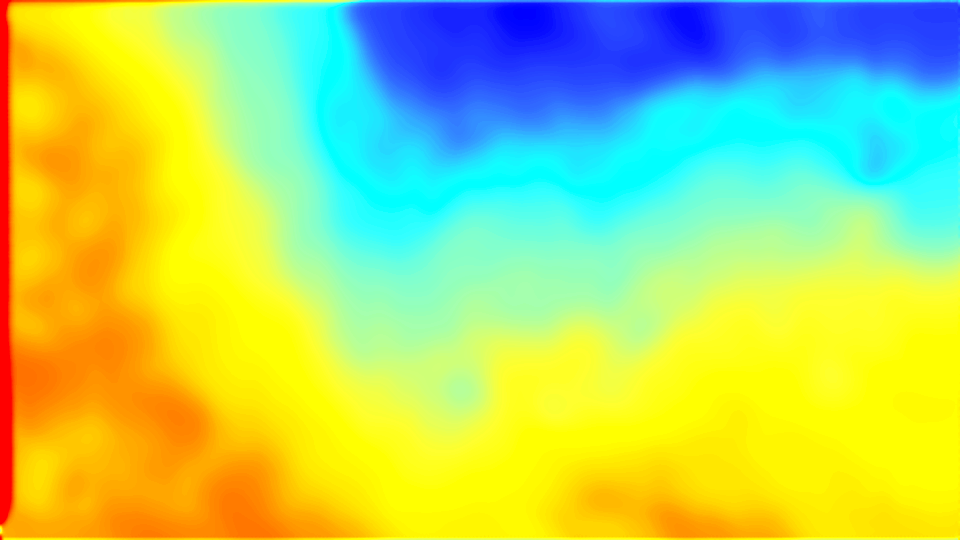}
			& 
\includegraphics[width=0.26\textwidth]{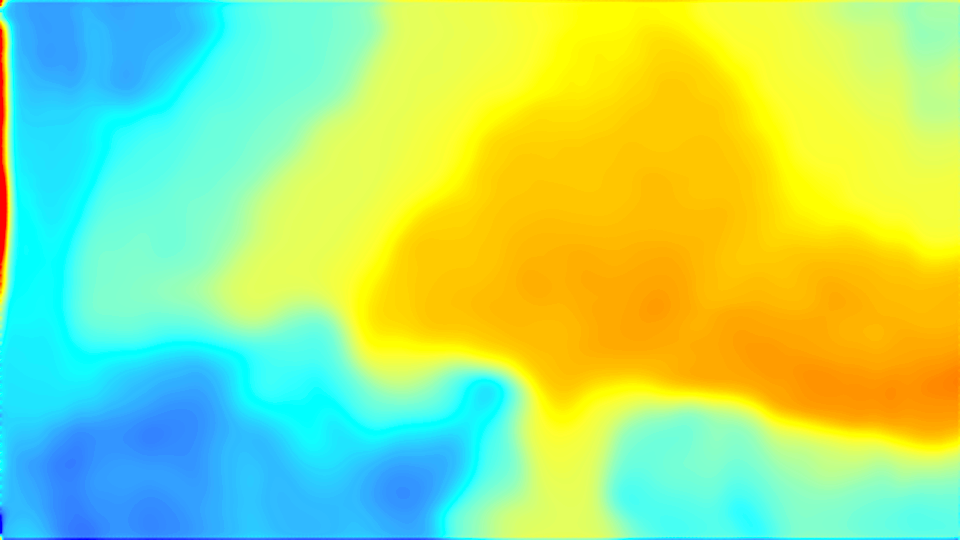}	
            &
\includegraphics[width=0.26\textwidth]{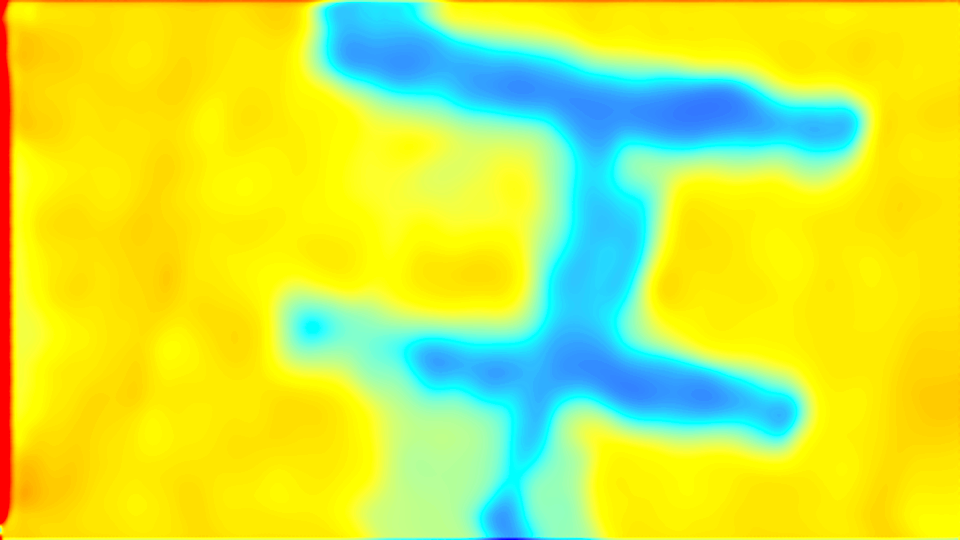}
\\   

\includegraphics[width=0.26\textwidth]{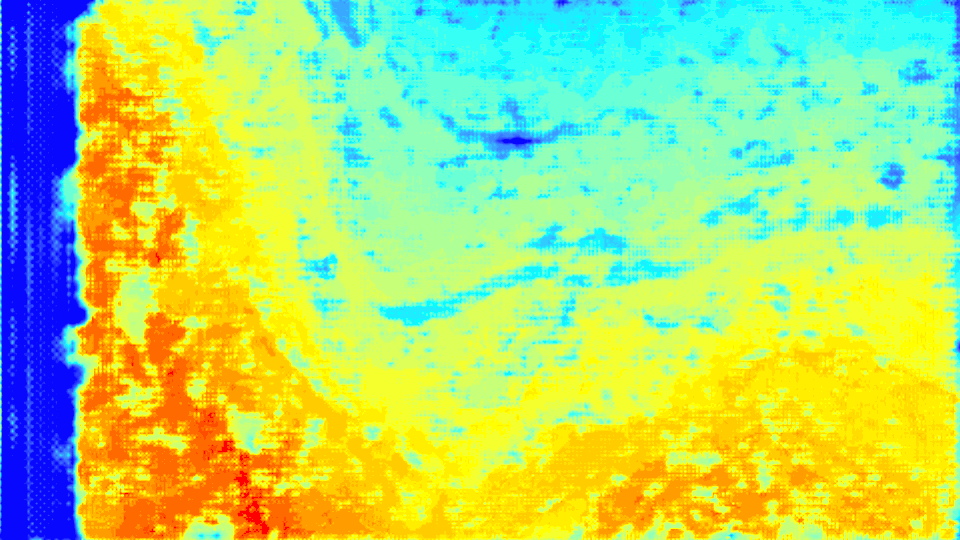}
			& 
\includegraphics[width=0.26\textwidth]{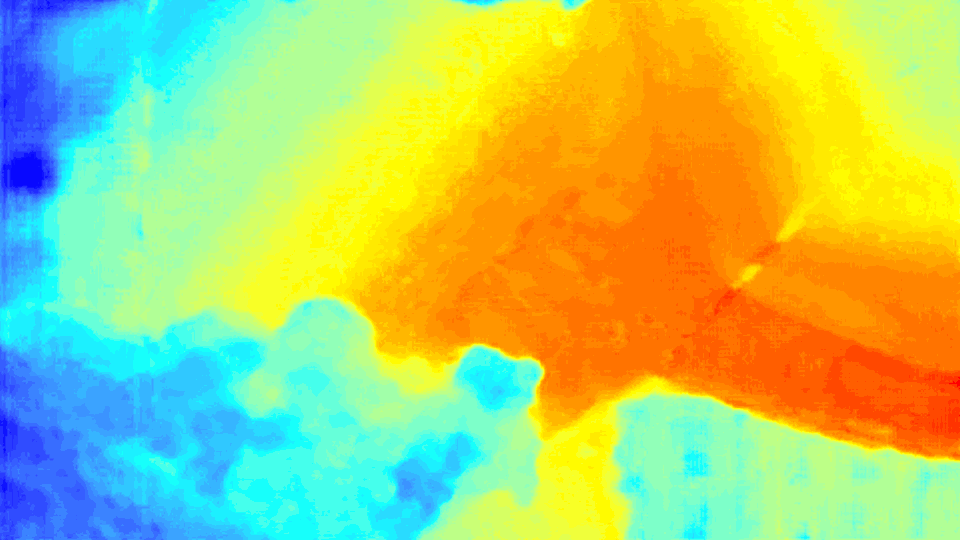}	
            &
\includegraphics[width=0.26\textwidth]{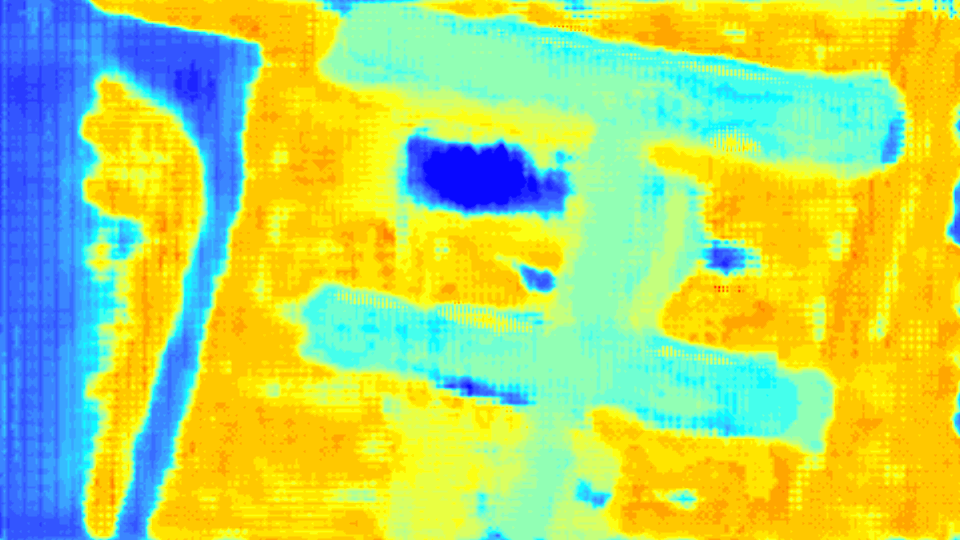}
\\   

\includegraphics[width=0.26\textwidth]{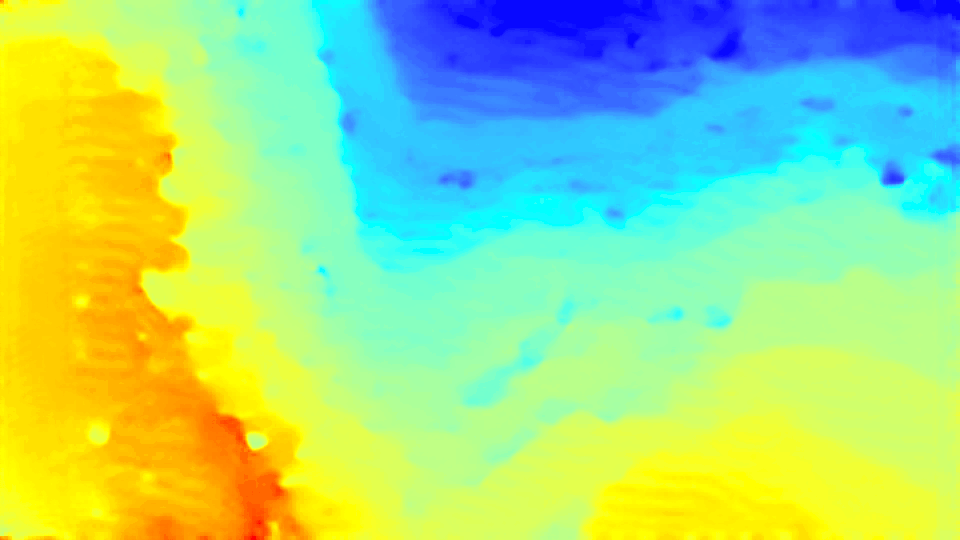}
			& 
\includegraphics[width=0.26\textwidth]{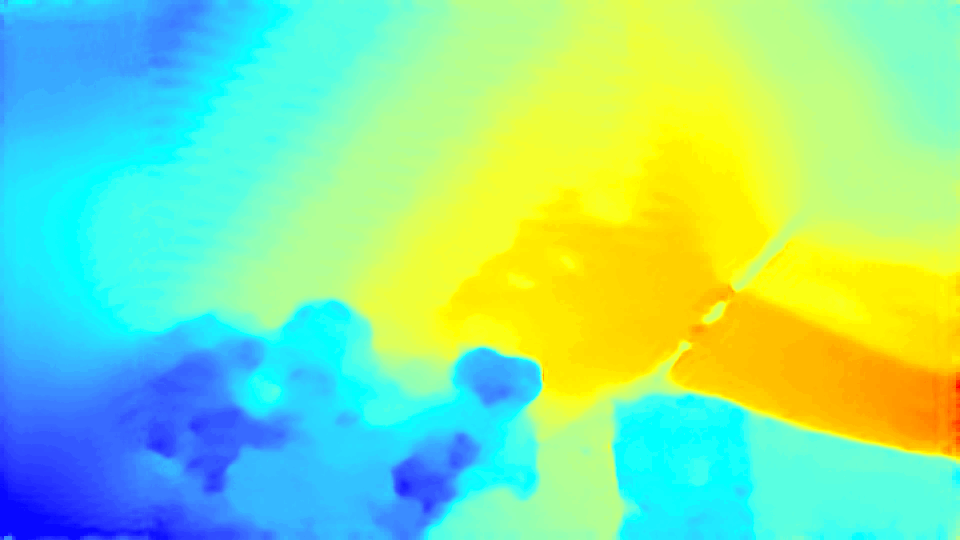}	
            &
\includegraphics[width=0.26\textwidth]{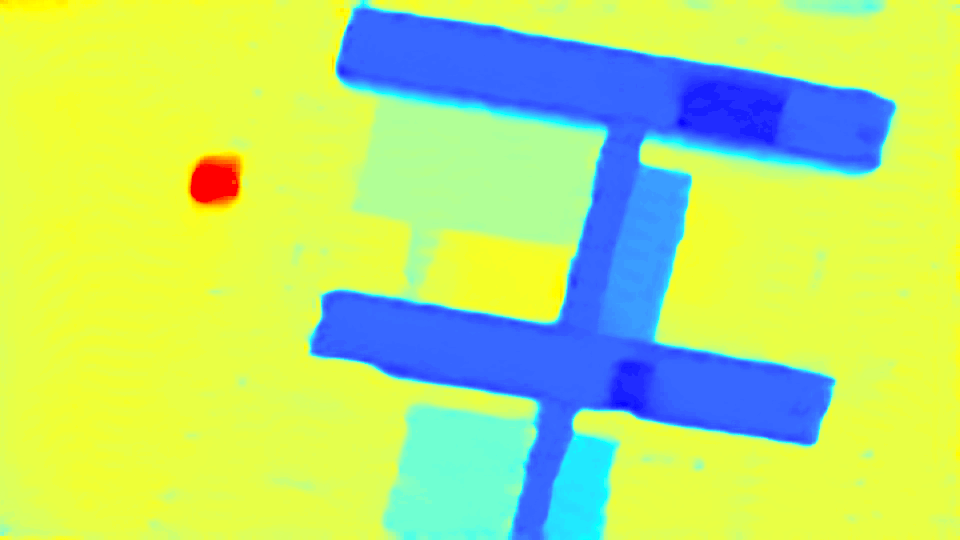}
\end{tabular} }  %

 \caption{Comparison of disparity maps estimated by different methods. From top to bottom: left image, ground truth disparity map, SGM, PSMNet, DSMNet, CFNet, RAFT-Stereo}
\label{tab:result}
\end{table*}

The disparity searching range increases as images resolution does. As shown in Tab. \ref{tab:samplesSynImages} and \ref{tab:samplesrealImages}, the disparity searching range of $1920 × 1080 px$ and 3840 × 2160 $px$ should be set to 768 $px$ and 960 $px$, causing an increase in computing memory usage. Since networks inferences are performed on a single GPU, most algorithms can run only at 960 × 540 $px$ because of memory constraints. Consequently, their predicted disparities are upsampled with bilinear interpolation in order to perform the comparison with higher resolution ground-truth maps, with predicted disparities scaled by the upsampling factor itself. In the bottom portion of Tab. \ref{tab:Syntest}, we list the evaluation results at 1920 × 1080 $px$ and 3840 × 2160 $px$. Due to the incompatibility of SGM and EAIStereo with drone imagery, we did not evaluate it on larger resolution. We can notice how all methods struggle at achieving good results at such high resolution, with RAFT-Stereo achieving best results. This result was expected because RAFT-Stereo achieves top-rank on Middlebury. Since all error metrics on low resolution images are still very far from those on existing benchmarks \cite{geiger2012we} \cite{mayer2016large} \cite{scharstein2014high}, we point out how they are still have large great challenge in geospatial UAV images due to the large presence of ground objects. By comparing the results of low and high resolution, we confirm that resolution is undoubtedly a challenge in our benchmark.

Whatsmore, comparing the test results on R, F and M subsets, the error of PSMNet in forest area is higher than that in residential area and mining area, whereas the other methods are just opposite. This suggests that DSMNet, CFNet, and RAFTStereo are more suitable to deal with disparity estimation of repetitive textures, like woods.

\subsection{Evaluation on Real subset}
\label{sec:real test}
In order to demonstrate the capacity of real subset in UAVStereo, we select the top-performing RAFT-Stereo network from previous evaluation, which is capable of handling huge disparity estimation. We conducted two experiments on the UAVStereo real subset: training the network directly with real training set and finetuning the synthetic pre-trained model using real data. In the latter experiment, we finetuned the pre-training model using only a quarter of the synthesized data. Take disparity range into account, we set the maximum disparity search range in the training stage of real data to 1920 $px$. Due to the large disparity value in the real scene, EPE, 30PE (30-pixel Error),and 100PE (100-pixel Error) were utilized to determine the inference error. In Tab. \ref{tab:Realtest}, we compared the metric errors among the pre-trained model on synthetic subset, the trained model on real subset and the finetuned model. 

It has been demonstrated that the real subset of UAVStereo can be used to train stereo matching models, though there is a large gap with the results on low resolution synthetic datasets. This may be related to the large disparity search range. In addition, we found that, although utilizing less data, the finetuned results were comparable to the training results. This confirms both our claims on the challenges in deep stereo networks as well as the significance of our dataset.

\section{Conclusions}
\label{sec:conc}
The quantity and quality of the dataset are critical to the performance of stereo matching algorithms. In this paper, a novel pipeline is proposed for generating image pairs and dense disparity for UAV scenes using images and point clouds. With the proposed pipeline, we have constructed the UAVStereo, a novel stereo dataset in UAV Scenarios - containing synthetic and real image pairs - featuring large disparity searching range and covering geospatial information, which is extremely challenging for the existing learning-based networks. Compared to available stereo datasets targeting autonomous driving, indoor, and aerial, UAVStereo is the first stereo matching dataset in UAV Scenarios including a large number of image pairs and dense label, which is supposed to speed up the process of 3D reconstruction.

UAVStereo dataset can represent to some extent the characteristics of UAV stereo matching data with large disparity search space, bigger possibility of ill-areas and more varied disparity distribution. Many experiments demonstrate that our dataset can be used for stereo matching of traditional algorithms and deep learning algorithms, and through experiments we find that deep learning-based methods have significant advantages over traditional algorithms and have great potential for development. Using a small amount of real data to finetune the pretrained model can achieve the comparable accuracy as the real data, and this strategy can effectively reduce the amount of real data required.

Our experiments show that UAVStereo unveils some of the most intriguing challenges in deep stereo and provides hints on promising research directions. In particular, followup work fostered by UAVStereo may be devoted to 1) investigating on the ability of deep models processing on large disparity searching range. 2) enhancing the capability of process geospatial data covering ground objects, which is great different from driving indoor scenes. 3) improve the generalization ability of the network between the synthetic domain and the real domain.

Therefore, we hope that UAVStereo holds the potential to improve future research in UAV scenario stereo matching and 3D reconstruction.

\bibliographystyle{ieeetr}
\bibliography{ref.bib}

          



        

\end{document}